\documentclass{article}

\PassOptionsToPackage{numbers, compress}{natbib}

\usepackage[main, final]{neurips_2026}
\usepackage[utf8]{inputenc} 
\usepackage[T1]{fontenc}    
\usepackage{hyperref}       
\usepackage{url}            
\usepackage{booktabs}       
\usepackage{amsfonts}       
\usepackage{nicefrac}       
\usepackage{microtype}      
\usepackage{xcolor}         
\usepackage{graphicx}
\usepackage{enumitem}
\usepackage{amsmath}
\usepackage{floatrow}
\usepackage{sidecap}
\usepackage{wrapfig}
\usepackage{natbib}
\usepackage{algorithm, algpseudocode, amsmath, amssymb, bm}
\usepackage{caption}
\usepackage{tabularx}
\usepackage{makecell}
\usepackage{authblk}
\usepackage{xltabular}
\usepackage{xspace}

\newcommand{\Nimbus}{\emph{Nimbus}\xspace}
\newcommand{\Bezier}{B\'{e}zier\xspace}

\title{The Marauder's Map: \Bezier Manifolds Reveal Hidden Surfaces for Model Merging and Ensembling}

\author[1, 3]{\textbf{Abhiram Iyer}}
\author[2, 3]{\\ \textbf{Mark T. Harnett}$^{\dagger}$}
\author[4]{\textbf{Sarthak Chandra}$^{\dagger}$}

\affil[1]{\footnotesize Department of Electrical Engineering and Computer Science, MIT, Cambridge, MA}
\affil[2]{\footnotesize Department of Brain and Cognitive Sciences, MIT, Cambridge, MA}
\affil[3]{\footnotesize McGovern Institute for Brain Research, MIT, Cambridge, MA}
\affil[4]{\footnotesize International Centre for Theoretical Sciences (ICTS-TIFR), Bengaluru, KA, India}
\affil[ ]{\footnotesize\tt \{abiyer, harnett\}@mit.edu, sarthak.chandra@icts.res.in}

\begin{document}

\maketitle

\begin{abstract}
Neural circuits are famously more robust than their deep network counterparts. This phenomenon is often explained through degeneracy, where structurally distinct configurations support the same function. However, real biological systems exhibit a more flexible form of degeneracy in which specialized circuits continually adapt towards new competencies through localized changes while preserving existing function. Deep learning studies degeneracy through mode connectivity, which connects same-objective networks by low-loss curves, sheets, or volumes. However, the field lacks an analogous counterpart to flexible degeneracy, which demands more than ordinary mode connectivity. Defining such an analogue requires asking whether many independently initialized models trained on distinct tasks can be connected by a high-dimensional region of competent multi-task solutions. To address this issue, we introduce \Nimbus, which learns a high-dimensional \Bezier manifold connecting $N$ specialists, going beyond the one-dimensional paths and two-dimensional sheets of prior mode connectivity work. A naive parameterization would scale combinatorially with $N$, but \Nimbus requires learning correction tensors of only $3$ interaction orders, showing that the connecting manifold has surprisingly low complexity yet supports multi-task competencies. The resulting manifold from \Nimbus forms a navigable ``Marauder's Map'' that connects independently trained specialist models through a shared geometry of competent solutions. Trained across vision and language domains, \Nimbus produces manifolds that are functionally and locally weight-flat: ensembles sampled from them degrade more gracefully than baselines under input and weight perturbations, with the largest gains in negative log-likelihood loss and expected calibration error. Thus, \Nimbus supplies a deep learning interpretation of flexible degeneracy, recasting model connectivity from a low-dimensional path between distant solutions into a high-dimensional volume of nearby ones.
\end{abstract}

\renewcommand{\thefootnote}{\fnsymbol{footnote}}
\footnotetext[2]{Equal supervision and co-corresponding authors.}
\renewcommand{\thefootnote}{\arabic{footnote}}
\setcounter{footnote}{0}

\section{Introduction}
\label{sec:introduction}

One of the key features of biological neural circuits is their robustness under synaptic noise. This defining characteristic has long been viewed through the lens of biological \emph{degeneracy}~\citep{tononi1999measures}, in which structurally distinct micro-configurations support the same macro-level function.
Analogously, deep learning studies a similar phenomenon but seeks to answer how distinct yet functionally similar solutions are \emph{connected} in high-dimensional space.
Mode connectivity shows that deep networks trained on the same objective can often be linked by low-loss curves or linear paths~\citep{garipov2018loss, draxler2018essentially, frankle2020linear}, with some work even extending to two-dimensional nonlinear sheets~\citep{mode_connectivity_bezier} and low-loss volumes~\citep{benton2021loss}.

\begin{wrapfigure}{r}{0.3\textwidth}
  \centering
  \vspace{-1pt}
  \includegraphics[width=\linewidth]{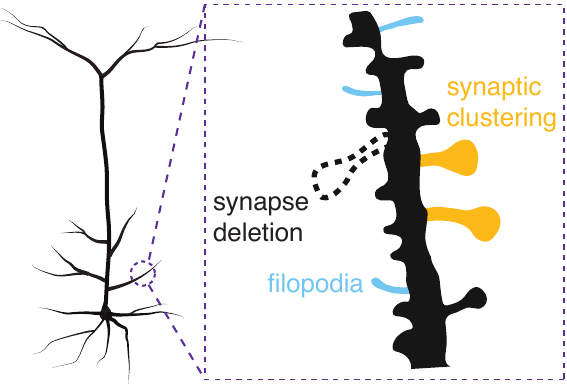}
  \caption{
  \textbf{Local degrees of freedom in biological learning.}
  Dendrites support synaptic clustering (yellow), deletion (dashed), and filopodia (blue) carrying silent synapses that can become active. These localized mechanisms of change let circuits learn without overwriting existing function.
  }
  \label{fig:dendrite}
  \vspace{-5pt}
\end{wrapfigure}

However, \emph{real} biological systems routinely undergo structural change as they learn new competencies. 
For example, dendritic spines grow and shrink, and can even transition a reservoir of silent synapses into active connections to directly alter synaptic strength.
Meanwhile, dendrites themselves undergo broader morphological remodeling, altering their overall length, branching complexity, and local signaling capacity. 
Together, these elements of structural change implement putative local learning mechanisms such as synaptic clustering or dendritic compartmentalization. 
These local degrees of freedom (Fig.~\ref{fig:dendrite}) are the means by which circuits can change without completely overwriting prior behavior~\citep{magee2020synaptic, otor2022dynamic, weber2016location, hedrick2022learning, hedrick2024local, vardalaki2025silent}.

\emph{Flexible} degeneracy is a more accurate description of biology, where specialized circuits can gradually adapt towards multiple new competencies through sequences of localized change while still preserving existing functionality.
The equivalent question in deep learning is clear.
Can a high-dimensional manifold connect \emph{many} independently initialized models that are trained on \emph{distinct} tasks while maintaining uniform task competency and retaining prior functionality?

We study exactly this. Given $N$ independently initialized models with a shared architecture but trained on different tasks, we learn a high-dimensional manifold connecting them, a setting that is stricter than ordinary mode connectivity as endpoints need not solve every task. Moreover, we will demonstrate that this manifold need not have a complex structure, but can instead be parameterized through a small number of tunable controls.
We propose \Nimbus: an \underline{N}-expert \underline{I}nterpolating \underline{M}anifold built as a \underline{B}ezier surface over an \underline{U}nderlying \underline{S}implex. The $N$ specialist models anchor the vertices of the simplex, and \Nimbus learns the $(N-1)$-dimensional manifold that fills its interior, supplying an infinite family of competent multi-task models.
Empirically, we will show that \Nimbus constructs functional manifolds between independently trained models in both vision and language.

Through this lens, \Nimbus operates under a richer premise in both mode connectivity \emph{and} ordinary model merging. Most merging methods~\citep{fisher_merging, jin2022dataless, ties_merging, zipit, yang2023adamerging} output a single model while assuming no access to task data. We instead operate in a distinct regime where task data is available and the goal is to learn the \emph{geometry} of the merge. Crucially, data here is not merely supervision for a single output model, but the foundation for building a high-dimensional manifold of models instead.

Once such a manifold has been constructed, we can also efficiently sample an arbitrarily large number of models for ensembling. Under input and weight perturbations, this ensemble model maintains strong accuracy while improving negative log-likelihood (NLL) loss and expected calibration error (ECE), metrics that gauge predictive quality and uncertainty more reliably than top-$K$ accuracy~\citep{guo2017calibration, ovadia2019can}. Without explicit flatness constraints, the process naturally yields a high-dimensional, functionally and locally weight-flat manifold of specialized models---a structural feature that conveniently aligns with established heuristics for out-of-distribution generalization.

\begin{figure}[h!]
\centering
    \includegraphics[width=0.86\textwidth]{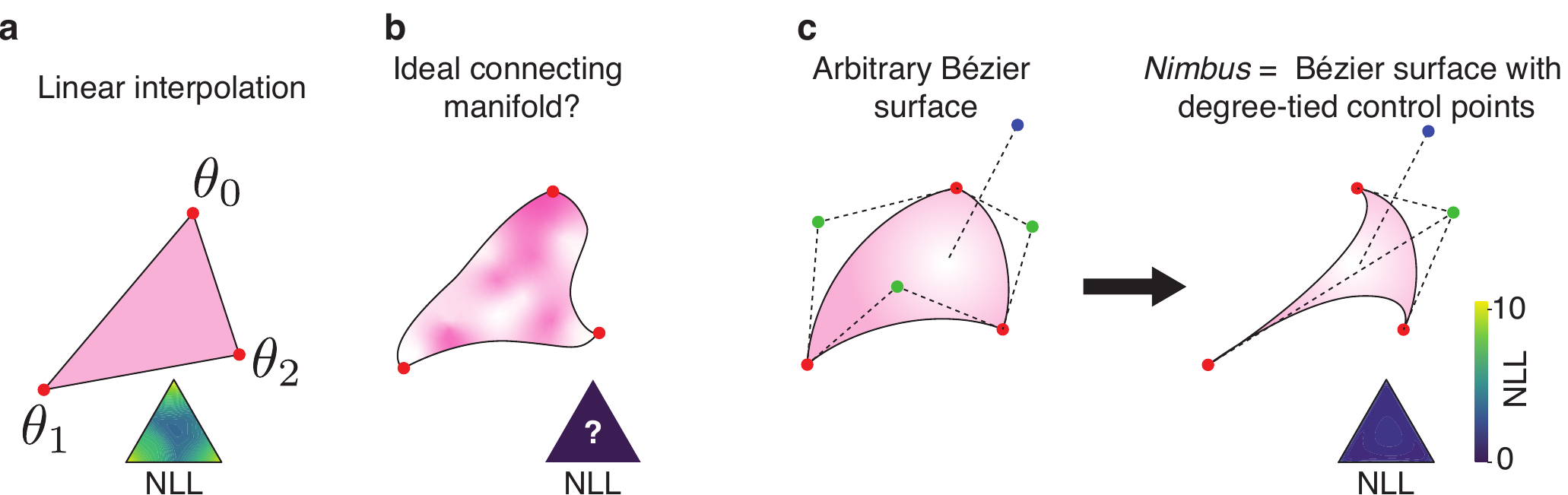}
    \caption{
    \textbf{\Nimbus constructs a low-loss, high-dimensional volume through a small number of corrections.}
    \textbf{a.} Constructing multi-task models through linear interpolation of independent task-trained models results in high NLL due to feature differences and neuron asymmetries.
    \textbf{b.} We hypothesize the existence of an ideal connecting manifold that can demonstrate low loss on the joint set of tasks in its entire volume.
    \textbf{c.} (\emph{left}) Could an arbitrary \Bezier manifold defined over a simplex satisfy this requirement? Even if it could, a naive parameterization would require combinatorially many correction terms, e.g., the green dots indicate learned quadratic corrections for every edge.
    (\emph{right}) \Nimbus demonstrates that such manifolds not only exist, but can be parameterized through only a small number of corrective terms shared per-degree.
    }
    \label{fig:schematic}
\end{figure}

\subsection{Related work}
\label{sec:related_work}

\paragraph{Mode connecting paths and surfaces.}
Mode connectivity asks whether trained networks are isolated minima or points in a connected low-loss set. 
\citet{gitrebasin} found that low-loss \emph{linear} paths between differently initialized same-objective models can only exist after neuron asymmetries are resolved via permutation.
\citet{garipov2018loss} and \citet{draxler2018essentially} found \emph{nonlinear} low-loss paths between modes by constructing simple curves, \citet{nguyen2019connected} gave theoretical conditions for connected sublevel sets, and \citet{zhou2023going} studied connectivity at the feature level. Beyond paths, \citet{benton2021loss} show that same-objective modes can lie in multi-dimensional low-loss simplicial complexes, and \citet{mode_connectivity_bezier} construct two-dimensional \Bezier surfaces connecting up to four networks. 
These are the closest geometric precedents to \Nimbus, but all assume \emph{same-objective} training and connect at most a handful of networks. \Nimbus drops both assumptions as \emph{many} models are \emph{independently} initialized and trained on \emph{disjoint} tasks.
Non-trivially, \Nimbus finds a high-dimensional mode-connected region with multi-task capabilities while also having specialist models at the boundaries of this region.

\paragraph{Model merging.}
Model merging combines trained models without retraining. Approaches include weight averaging in a compatible basin~\citep{wortsman2022model}, curvature-weighted combination (Fisher Merging~\citep{fisher_merging}), task-vector composition over a shared pre-trained base (Task Arithmetic and TIES Merging~\citep{ilharco2022editing, ties_merging}), feature-matched merging across specialist networks (ZipIt!~\citep{zipit}), permutation-based merging of same-objective models (Git Re-Basin~\citep{gitrebasin}), and data-less or adaptive variants~\citep{jin2022dataless, yang2023adamerging}, with recent LLM work absorbing capabilities from homologous specialists~\citep{yu2024language}. 
All collapse $N$ specialists into a single model and typically assume a shared pre-trained base or fine-tuning lineage. Feature- and permutation-matching variants further require near-identical tasks, or merge only part of the network and leave per-specialist heads that scale linearly with $N$.
\Nimbus instead preserves specialists exactly and adds only a constant number of learnable corrections to build a continuum of multi-task models between them, a geometry that can be traversed to tune task mixtures, sampled for ensembling, and stress-tested under perturbations.

\paragraph{Single-model solutions for robustness and multi-task training.}
A separate line of work improves robustness or mitigates task interference at the level of a single solution. PCGrad addresses interference during joint training by projecting conflicting gradients onto orthogonal subspaces~\citep{pcgrad}. Deep ensembles remain strong baselines under distribution shifts~\citep{lakshminarayanan2017simple, ovadia2019can} but require many independent models. Stochastic weight averaging (SWA) averages along an SGD trajectory to find wider optima~\citep{swa}, and Adaptive Sharpness-Aware Minimization (ASAM) explicitly optimizes for weight flatness~\citep{asam}. 
\Nimbus presents a continuum with a closed functional form, so ensembling can operate over an arbitrarily large set of models. Even without explicit functional and weight flatness penalties, its ensemble outperforms these baselines under perturbations, suggesting that the learned manifold is naturally well-conditioned.

\paragraph{Contributions.} We make the following key contributions:

\begin{itemize}[leftmargin=*]
    \item \textbf{A high-dimensional manifold connecting independently trained specialists.} Prior mode-connectivity work links models through low-loss paths and surfaces only when they share the same objective. \Nimbus vastly generalizes this paradigm by establishing an $(N-1)$-dimensional low-loss manifold across $N$ specialists trained with \emph{distinct} initializations on \emph{distinct} training tasks.

    \item \textbf{Connectivity manifold has low intrinsic complexity.} The high-dimensional manifold is parameterized as a linear mixture of the $N$ frozen specialists plus a small number of additive correction tensors that capture how groups of specialists interact. A natural parameterization would assign one correction per face of the simplex and would grow exponentially in $N$. Non-trivially, \Nimbus finds that just $3$ interaction orders are sufficient to build competent manifolds, keeping the construction tractable at scale.

    \item \textbf{Functionally and locally weight-flat manifolds.} When \Nimbus is trained across vision and language domains, sampling from its manifold yields ensembles that degrade more gracefully than baselines under input and weight perturbations, with the largest gains in NLL and ECE, metrics that best capture predictive-distribution quality.

    \item \textbf{A deep learning instantiation of flexible neural degeneracy.}
    Just as a neural circuit uses localized mechanisms to acquire new competencies in high-dimensional space without catastrophic overwriting (i.e., flexible degeneracy), a specialist can smoothly accumulate new task competencies on \Nimbus's manifold while small synaptic perturbations on the same manifold yield functionally equivalent yet differently configured models. Our work fundamentally unifies neural degeneracy with mode connectivity and model merging, consequently recasting degeneracy from low-dimensional surfaces to high-dimensional volumes.
\end{itemize}

\section{How do we construct a \Bezier manifold between models?}
\label{sec:method}

\subsection{Preliminaries}

We are given $N$ models $\{f_{\theta_i}\}_{i=1}^N$ that share a single
architecture $f_\theta$ and have been trained independently on different tasks
$\{\mathcal{T}_i\}_{i=1}^N$ with independent arbitrary initializations. 
Our goal is to construct a single \emph{parameter map} $\theta : \Delta^{N-1} \to \mathbb{R}^{|\theta|}$, where $\Delta^{N-1}$ is the $(N-1)$-simplex ($\Delta^{N-1} = \{ \bm{w} \in \mathbb{R}^N : w_i \geq 0, \sum_{i=1}^{N} w_i = 1 \}$), such that $\theta(\bm{w})$ at any mixing weight $\bm{w} \in \Delta^{N-1}$ yields a model that performs well on the combination of tasks corresponding to the nonzero entries of $\bm{w}$. Specifically, this demands that the vertices $\bm{e}_i$ of the simplex $\Delta$ are recovered exactly, i.e., $\theta(\bm{w} = \bm{e}_i) = \theta_i$, and that interior $\bm{w}$ yields models that behave like genuine multi-task models rather than naive parameter averages.

Model soups and linear mode connectivity~\citep{wortsman2022model, frankle2020linear} typically employ a linear combination of weights (i.e., the parameter map $\theta(\bm{w}) = \sum_i w_i \theta_i$). However, when tasks are disjointly defined and models are free to start from arbitrary initializations, linear combinations are insufficient due to conflicting features and neuron asymmetries (Fig.~\ref{fig:schematic}a). Given this practical constraint, we seek to construct parameter maps through nonlinear combinations of model weights to answer whether such a low-loss manifold spanning specialist models can be constructed automatically (Fig.~\ref{fig:schematic}b).

\subsection{\Bezier manifolds}

We consider an $N$-variate, $d$-degree \Bezier surface as our natural starting point. Such a manifold is typically written in the form of:

\vspace{-10pt}
\begin{equation}
    \textstyle
    \theta(\bm{w}) = \sum_{|\bm{\alpha}| = d} b_{\bm{\alpha}}^{d}(\bm{w}) P_{\bm{\alpha}}, \quad b_{\bm{\alpha}}^{d} = \frac{d!}{\alpha_1! \alpha_2! ~\text{...}~ \alpha_N!} \prod_{i=1}^{N} w_i^{\alpha_i}
\label{eq:bezier_basic}
\end{equation}
\vspace{-12pt}

where $\bm{\alpha}$ is a multi-index that iterates over all interaction terms on the $(N-1)$-simplex, $b_{\bm{\alpha}}^{d}$ are the Bernstein basis polynomials, and $P_{\bm{\alpha}}$ are the \Bezier (or Bernstein) coefficients (conventionally referred to in mathematical literature as ``control points''). We note that when $\bm{\alpha} = d \bm{e}_i$ for unit basis vector $\bm{e}_i \in \mathbb{R}^{N}$, $P_{\bm{\alpha}} = \theta_i$ is the original $i$-th vertex model exactly. We can reinterpret Eq.~\ref{eq:bezier_basic} as a linear combination of vertex models plus additive correction terms applied to each face of the simplex (see Supp. Sec.~\ref{sec:proof} for an expanded proof).

\vspace{-10pt}
\begin{equation}
    \textstyle
    \theta(\bm{w}) 
    = \sum_{i=1}^{N} w_i \theta_i 
    + \sum_{|\bm{\alpha}| = d} b_{\bm{\alpha}}^d(\bm{w})\,\tilde{P}_{\bm{\alpha}},
    \qquad 
    \tilde{P}_{\bm{\alpha}} 
    := P_{\bm{\alpha}} - \sum_{i=1}^{N} \frac{\alpha_i}{d}\theta_i .
    \label{eq:bezier_linear_interp_plus_bump}
\end{equation}
\vspace{-12pt}

Here, $\tilde{P}_{\bm{\alpha}}$ is the residual Bernstein coefficient after subtracting the affine interpolation of the vertex models. If $N=3$, for instance, these corrections are added to the three pairwise edges and the single $2$-face of the $\{\theta_0, \theta_1, \theta_2\}$ triangle.

\subsection{Learnable corrections on a \Bezier manifold}

Our method falls directly out of Eq.~\ref{eq:bezier_linear_interp_plus_bump} given a specific choice of $b_{\bm{\alpha}}^{d}$ and learnable $\tilde{P}_{\bm{\alpha}}$, where $|\tilde{P}_{\bm{\alpha}}| = |\theta|$. We use $\tilde{b}_{\bm{\alpha}}^{d} = \frac{ b_{\bm{\alpha}}^{d}(\bm{w}) }{ b_{\bm{\alpha}}^{d}(\frac{\bm{\alpha}}{d}) } = \prod_i \left( \frac{d w_i}{\alpha_i} \right)^{\alpha_i}$, which is normalized by the maximum value that the Bernstein basis polynomial can take under the function $\bm{w} \mapsto \prod_i (w_i)^{\alpha_i}$.

We model $\tilde{P}_{\bm{\alpha}}$ by trying to strike the right balance between maximizing expressivity and minimizing space complexity. If we take Eq.~\ref{eq:bezier_linear_interp_plus_bump} as is, then the number of $\tilde{P}_{\bm{\alpha}}$ parameters is exorbitantly high and scales as $\mathcal{O} \left( |\theta| \cdot \binom{N + d - 1}{d} \right) = \mathcal{O} \left( |\theta| \cdot {d^{N-1}} \right)$. 
First, we restrict $\bm{\alpha} \in \{0, 1\}^N$ to shrink the per-degree basis from $\binom{N+d-1}{d}$ to $\binom{N}{d}$, eliminating self- and cross-interaction terms (e.g., $w_i^2$ and $w_i^2 w_j$) that do not add considerable expressivity beyond the linear term (i.e., $\sum_{i} w_i \theta_i$) and multilinear terms (e.g., $w_iw_j$). As a consequence, the highest interaction order is capped at $N$. Second, given this ceiling, we choose which interaction orders we require from $\{2,...,N\}$ instead of operating with a fixed degree $d$. Third, we share $\tilde{P}_{\bm{\alpha}}$ for all interaction terms within a single degree. These decisions not only collapse the parameter count to $\mathcal{O}\left( |\theta| \cdot |\mathcal{D}| \right)$, independent of $d$, but also impose a permutation symmetry across vertex models at each interaction order (Fig.~\ref{fig:schematic}c). This symmetry crucially encodes the prior that the \emph{order} of an interaction matters but the \emph{identity} of which vertices participate at that order does not.
Further discussion of these design choices is included in Supp. Sec.~\ref{sec:algorithm_ablation}.

Taken together, we turn Eq.~\ref{eq:bezier_linear_interp_plus_bump} into:

\vspace{-10pt}
\begin{equation}
    \textstyle
    \theta(\bm{w}) = \sum_{i=1}^{N} w_i \theta_i + \sum_{r \in \mathcal{D}} ~\sum_{\substack{\bm{\alpha} \in \{0,1\}^N \\ |\bm{\alpha}| = r}}
    \tilde{b}_{\bm{\alpha}}^r(\bm{w})\, \tilde{P}_r.
    \label{eq:bezier_full}
\end{equation}
\vspace{-12pt}

Geometrically, this construction iterates through faces of the simplex by dimension: each $r \in \mathcal{D}$ contributes a correction supported on the $(r-1)$-faces of $\Delta^{N-1}$. Therefore, $\mathcal{D}$ ultimately selects which face dimensions carry learnable structure. For an $N$-vertex \Bezier manifold, we found that $\mathcal{D} = \{2, N-1, N\}$ was sufficient to capture the salient interactions between all vertices as this included both the dominant and highest-order interactions. Because $|\mathcal{D}| \leq 3$, the parameter count ultimately reduces to $\mathcal{O} \left( |\theta| \right)$, independent of both $d$ and $N$. In-depth ablations of each $r \in \mathcal{D}$ are included in Supp. Sec.~\ref{sec:interaction_ablation}.

\subsection{Algorithm}
\label{sec:algorithm}

In practice, we augment the parameter map of~\eqref{eq:bezier_full} with one learnable scalar gate per degree, $g_r \in \mathbb{R}$ for $r \in \mathcal{D}$, that multiplicatively rescales each correction:

\vspace{-10pt}
\begin{equation}
    \textstyle
    \theta(\bm{w}) = \sum_{i=1}^{N} w_i\, \theta_i
    + \sum_{r \in \mathcal{D}} e^{g_r} \sum_{\substack{\bm{\alpha} \in \{0,1\}^N \\ |\bm{\alpha}| = r}}
    \tilde{b}_{\bm{\alpha}}^r(\bm{w})\, \tilde{P}_r.
    \label{eq:nimbus}
\end{equation}
\vspace{-12pt}

Gates are initialized to $0$ (so $e^{g_r} = 1$) and provide a small, cheap handle for calibrating each correction's magnitude once the correction for a degree has been learned. We train \Nimbus in two phases: Phase~1 fits the correction terms $\{\tilde{P}_r\}$ while Phase~2 freezes these corrections and tunes only the gates. The original $\{ \theta_i \}$ always remain frozen. The full pseudocode is outlined in Supp. Sec.~\ref{sec:pseudocode}, but we include a summary here:

\paragraph{Phase~1: low-to-high degree hierarchical training.} 
We take inspiration from~\citet{kobayashi2019simplex}, which proposes fitting optimization problems on a simplex from lower faces to higher faces. We therefore fit our correction terms in increasing order of degree. Additionally, when training a particular correction, the gates of all previous degrees are unfrozen for joint refinement. 
Each step draws one minibatch per task and samples a small subset of degree-$r$ faces. On each face, we sample a set of $\bm{w}$ from a $50:50$ mixture of $\mathrm{Dirichlet}(r \cdot \mathbf{1}_r)$ and $\mathrm{Dirichlet}(\mathbf{1}_r)$. The first term puts its mass at each face's center (where the interaction we are correcting has the highest mass), while the second term spreads samples across the rest of the face, including near its lower-dimensional edges. 
Weighting the $\mathrm{Dirichlet}$ terms equally trains each degree across its entire face, ensuring correction tensors neither overfit the face center nor ignore the boundaries where they must blend smoothly into fixed lower-order edge residuals.
On each sampled $\theta\left({\bm{w}}\right)$, we minimize a cross-entropy loss on the combined labels from the tasks on the face, plus a KL-divergence to each task's vertex model evaluated on that task's own data.

\paragraph{Phase~2: gate calibration.}
We freeze all correction terms and only train the gates $\{g_r\}$. We sample a set of $\bm{w}$ uniformly across degrees $r \in \mathcal{D}$ and uniformly among all faces within each degree. On the pooled data of all tasks across the sampled faces, we minimize the mean (i) cross-entropy of the averaged logits, (ii) NLL of the averaged probabilities, and (iii) KL-divergence to each task's vertex model evaluated on that task's own data.
We ablate these design choices in Supp. Sec.~\ref{sec:algorithm_ablation} and conclude that \Nimbus's shared-by-degree corrections (covering pairwise and highest-order interactions), KL distillation losses, and Phase~2 training all contribute to a uniformly low-loss manifold.

\section{Experimental results}
\label{sec:results}

\subsection{Tasks and architectures}

We evaluate \Nimbus across both vision and language modalities. 

\paragraph{ConvNeXt.} For vision tasks, we train separate, randomly initialized ConvNeXt models~\citep{liu2022convnet} on subsets of CIFAR-100~\citep{cifar100}, where each task consists of $10$ disjoint classes selected under a fixed random permutation across $10N$ total classes ($N \in \{3, 5, 7, 10\}$). All models have a $10N$-node output head. The $|\mathcal{D}| \leq 3$ correction tensors $\tilde{P}_r$ use the same architecture.

\paragraph{ViT-B/32.} We additionally consider the ViT-B/32 vision encoder from OpenAI CLIP~\citep{radford2021clip}, a large-scale vision transformer ($32 \times 32$ patch size, $224 \times 224$ input image size, and approximately $87.5$M parameters) that was jointly pre-trained with a transformer text encoder on roughly $400$M image-text pairs. We fine-tune separate, pre-trained ViT-B/32 models via LoRA~\citep{lora} (rank$=64$, applied only to the attention projections and MLPs) on $N = 8$ vision tasks discussed in Task Arithmetic~\citep{ilharco2022editing}, which include MNIST (digits), SVHN (digits), GTSRB (traffic signs), DTD (perceptual textures), EuroSAT (satellite imagery), Cars, RESISC45 (satellite imagery), and SUN397 (naturalistic scenes). The $|\mathcal{D}| = 3$ correction tensors $\tilde{P}_r$ use the same architecture and are optimized similarly with LoRA under the same setting.
To perform zero-shot classification, we replace the final linear layer with class embeddings generated by CLIP's text encoder using task-specific prompt templates. Given an input image, the vision encoder extracts a feature vector, and the model assigns the class label whose normalized text embedding ($512$-dim) yields the highest cosine similarity with the image feature.

\paragraph{nanoGPT.} For language, we train separate, randomly initialized nanoGPT models~\citep{Karpathy2022} on $N = 7$ different pieces of text, which include Early Modern English (Shakespeare, the KJV, Milton, Jonson, and other in-period authors) from Project Gutenberg~\citep{gutenberg_website}, WikiText-103~\citep{wikitext}, TinyStories~\citep{tinystories}, Python source code (CPython stdlib~\citep{cpython} and CodeParrot~\citep{codeparrot}), arXiv papers~\citep{he2019arxiv}, U.S.\ legal text (Pile of Law~\citep{henderson2022pileoflaw}, English-only subsets), and French/German literature (Project Gutenberg~\citep{gutenberg_website}). All domains are tokenized with a shared $2048$-token BPE vocabulary ($2048$-node output head) fit on a concatenation of the corpora. The $|\mathcal{D}| = 3$ correction tensors $\tilde{P}_r$ use the same architecture.

Hyperparameters, experimental settings, and code are given in Supp. Sec.~\ref{sec:hyperparams}.

\subsection{\Nimbus admits flat functional manifolds}

\begin{figure}[h!]
\centering
    \includegraphics[width=1.0\textwidth]{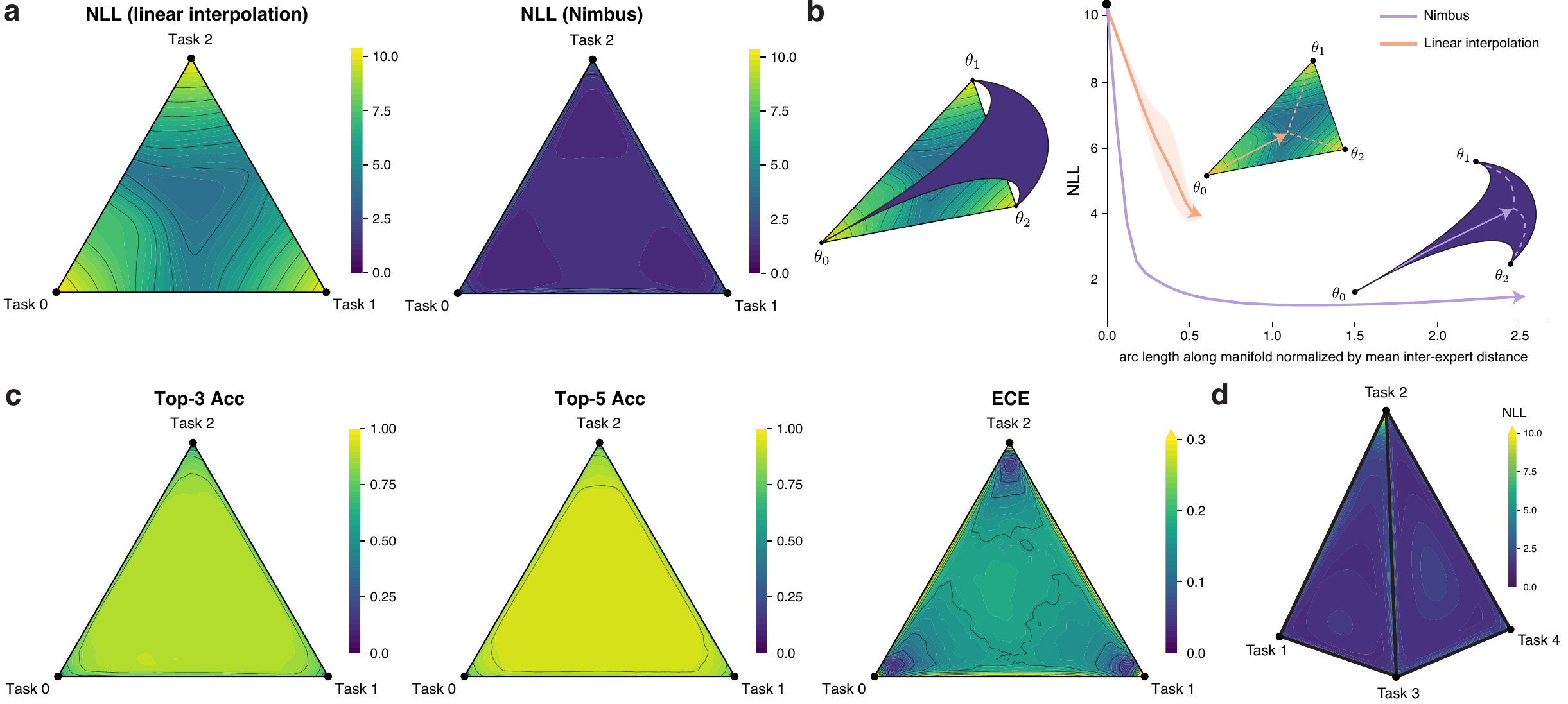}
    \caption{
    \textbf{The \Bezier manifold induced by \Nimbus is flat in loss.}
    \textbf{a.} Each point in the triangle heatmap corresponds to a mixing weight $\bm{w} \in \Delta^2$. For ConvNeXt ($N = 3$, $\mathcal{D} = \{2, 3\}$), linear connectivity between $3$ disjoint models crosses many high-loss barriers, while \Nimbus is uniformly low in NLL loss (lower is better). NLL $\approx 3.4$ is chance. 
    \textbf{b.} We can compare the linear paths towards the vertex average and the paths towards \Nimbus's centroid starting from any vertex. After moving roughly half an inter-expert distance, linear interpolation produces poorly functioning models while the path along \Nimbus's manifold yields access to an infinite set of highly competent solutions.
    \textbf{c.} Top-3 and top-5 accuracy (higher is better) stay uniformly high across the interior, while ECE (lower is better) stays uniformly low, indicating that any convex combination of specialists (plus learned corrections) yields a competent merged multi-task model. 
    \textbf{d.} Even when \Nimbus is optimized over $N = 4$ specialists, the low-loss $\Delta^3$ tetrahedron is bounded by lower-dimensional surfaces, each of which effectively merges subsets of specialist models. We visualize here two unique sets of $N = 3$ tasks as two faces of the tetrahedron. NLL $\approx 3.7$ is chance.
    }
    \label{fig:manifolds}
\end{figure}

We begin by considering \Nimbus instantiated as ConvNeXt ($N = 3$, $\mathcal{D} = \{2, 3\}$) and visualize the loss manifold induced by \Nimbus over $\Delta^2$, where vertices of the triangle are given by the $3$ task specialists and mixing weight $\bm{w} \in \Delta^2$ refers to points in this triangle. Fig.~\ref{fig:manifolds}a demonstrates that naive linear interpolation across these specialists encounters high-loss barriers, while \Nimbus is uniformly low in NLL loss despite models starting from different initializations and learning entirely different tasks. As $\bm{w}$ approaches a particular vertex, we also recover that vertex and its task performance through our parameter map $\theta(\bm{w})$. 

Remarkably, building manifolds through \Nimbus demonstrates that it suffices to move a small synaptic distance from any vertex towards the centroid of the manifold in order to access an \emph{infinite} set of competent solutions. We can validate this theory by tracing and averaging the loss along $N$ different ``rays'', each starting from a vertex and moving toward the manifold centroid (see Fig.~\ref{fig:manifolds}b). Distance along each ray is measured in Euclidean parameter space and normalized by the mean pairwise inter-expert distance. While the multi-task NLL loss near any one expert is $\approx 10.0$ ($\approx 3.4$ is chance), moving roughly half an inter-expert distance along \Nimbus's manifold collapses the loss to $\approx 1.0$. The average loss across these rays then stays in a broad, flat low-loss basin all the way across the interior. By contrast, the linear weight-average crosses many high-loss barriers, reaching a multi-task NLL of $\approx 4.0$ near the interpolation centroid. 

Fig.~\ref{fig:manifolds}c demonstrates functional flatness through uniformly high top-3 and top-5 accuracy and uniformly low ECE, indicating that any convex combination of specialists (plus learned corrections) yields a competent merged multi-task model. Even when \Nimbus is optimized over $N = 4$ specialists, the low-loss volume visualized as a $\Delta^3$ tetrahedron is bounded by lower-dimensional surfaces, each of which effectively merges subsets of specialist models (see Fig.~\ref{fig:manifolds}d). Thus, faces of the tetrahedron correspond to unique sets of $N = 3$ tasks that are also uniformly low in NLL loss. Despite a simplex face count that grows exponentially (here, $\binom{4}{4} + \binom{4}{3} + \binom{4}{2} + \binom{4}{1} = 15$), we conclude that a small number of correction tensors shared per order ($|\mathcal{D}| = |\{2, 3, 4\}| = 3$) suffices to populate the simplex with competent models. 

\subsection{Ensembling with \Nimbus}

\Nimbus's manifold admits a continuum of models, so we can use it for ensembling under both logit and softmax averaging. Under both rules, we fix a sampling budget split approximately equally across $r \in \mathcal{D}$, then uniformly across the faces within each degree, with weights drawn from a Dirichlet over the vertices of each sampled face.

\begin{figure}[h!]
\centering
    \includegraphics[width=0.80\textwidth]{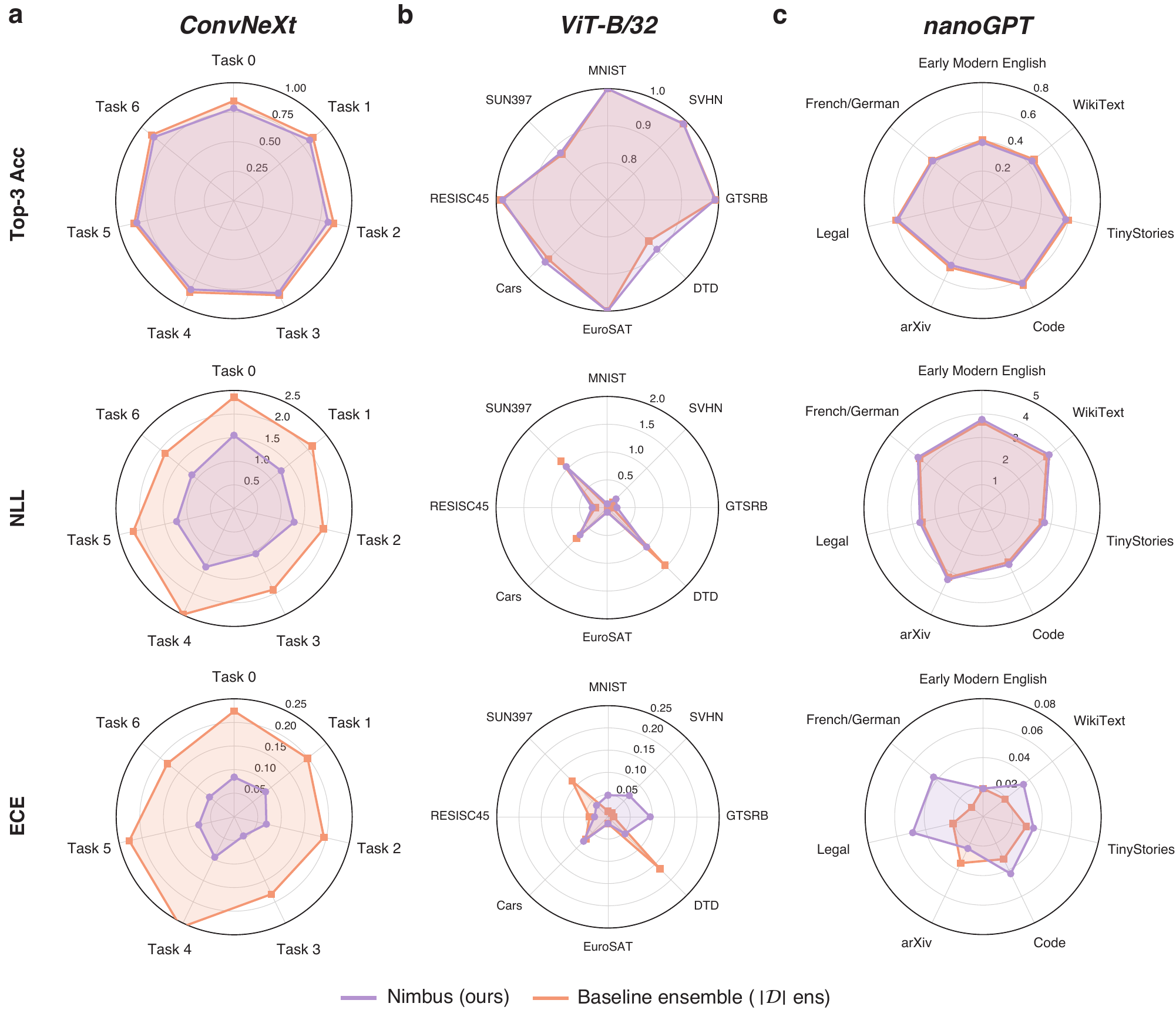}
    \caption{
    \textbf{Per-task metrics for ensembling on \Nimbus's manifold versus an ensemble of jointly trained models.}
    \textbf{a.} ConvNeXt ($N = 7$), \textbf{b.} ViT-B/32 ($N = 8$), and \textbf{c.} nanoGPT ($N = 7$) per-task metrics such as top-3 accuracy (higher is better), NLL (lower is better), and ECE (lower is better) under logit ensembling for \Nimbus (purple) versus a baseline ensemble of jointly trained models (orange). \Nimbus achieves lower NLL and comparable or higher top-3 accuracy than the baseline on every task in all settings and is well-calibrated under logit ensembling.
    }
    \label{fig:radar}
\end{figure}

Fig.~\ref{fig:radar} compares \Nimbus against a baseline ensemble of $|\mathcal{D}|$ arbitrarily initialized jointly trained models, each trained over all tasks. For simplicity, we report logit ensembling here and include full results (softmax ensembling, raw parameter averaging) in Supp. Sec.~\ref{sec:full_experiments} along with further discussion on the differences between the ensembling rules. On logit ensembling, \Nimbus crucially achieves lower NLL on every task across ConvNeXt ($N = 7$), ViT-B/32 ($N = 8$), and nanoGPT ($N = 7$), and is comparably well-calibrated. 

\subsection{\Nimbus demonstrates robustness under input and weight perturbations}

Next, we evaluate \Nimbus under distribution shifts simulated by input and weight perturbations, which probe functional and synaptic weight flatness, respectively.  

We compare against three baselines matched to \Nimbus on both total learnable parameters and total training compute, under progressively more conservative accounting of what a practitioner is responsible for (see Supp. Sec.~\ref{sec:compute_matching} for a rigorous treatment of this topic). The first is an ensemble of $|\mathcal{D}|$ models fine-tuned jointly on all tasks, splitting \Nimbus's budget evenly among them. The second charges \Nimbus for the cost of training its $N$ vertices as well, expanding the ensemble to $|\mathcal{D}| + N$ models at the same per-model budget (effectively $\frac{N}{|\mathcal{D}|} \times$ \emph{more} compute). The third assumes the practitioner does not want to ensemble at all and thus trains a single joint model with $2 \times$ the compute and $2 \times$ the parameters of \Nimbus (effectively $4 \times$ the compute). Additional baselines such as linear interpolation, ASAM, SWA, PCGrad, Fisher Merging, TIES Merging, and Task Arithmetic~\citep{asam, swa, pcgrad, fisher_merging, ties_merging, ilharco2022editing} are included with the full results in Supp. Sec.~\ref{sec:full_experiments}; \Nimbus outperforms each substantially.

\begin{figure}[h!]
\centering
    \includegraphics[width=0.85\textwidth]{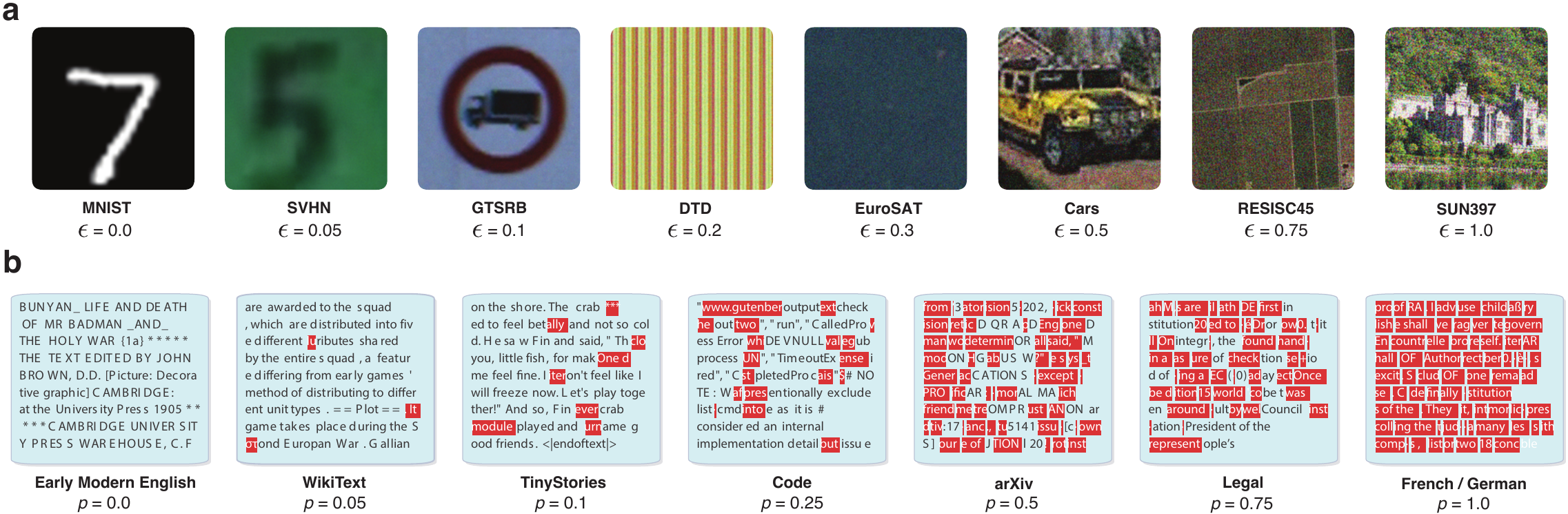}
    \caption{
    \textbf{Input corruption for vision and language experiments.}
    \textbf{a.} Sample image tasks for ViT-B/32 experiments. Gaussian noise of magnitude $\varepsilon \in \{0.0, 0.05, 0.1, 0.2, 0.3, 0.5, 0.75, 1.0\}$ relative to each image's $\ell_2$-norm is added.
    \textbf{b.} Sample text tasks for nanoGPT experiments. Input tokens are independently replaced with a uniform random token from the BPE vocabulary with probability $p \in \{ 0.0, 0.05, 0.1, 0.25, 0.5, 0.75, 1.0 \}$. Corrupted tokens are highlighted in red.
    }
    \label{fig:input_perturbation_visualization}
\end{figure}

\paragraph{Input perturbations.} 
Within the vision modality, we perturb input images with additive Gaussian noise of magnitude $\varepsilon$ relative to the $\ell_2$-norm of each test image (see Fig.~\ref{fig:input_perturbation_visualization}a). 
Within the language modality, we randomly substitute each input token with probability $p \in [0, 1]$ (see Fig.~\ref{fig:input_perturbation_visualization}b).

\begin{figure}[h!]
\centering
    \includegraphics[width=0.80\textwidth]{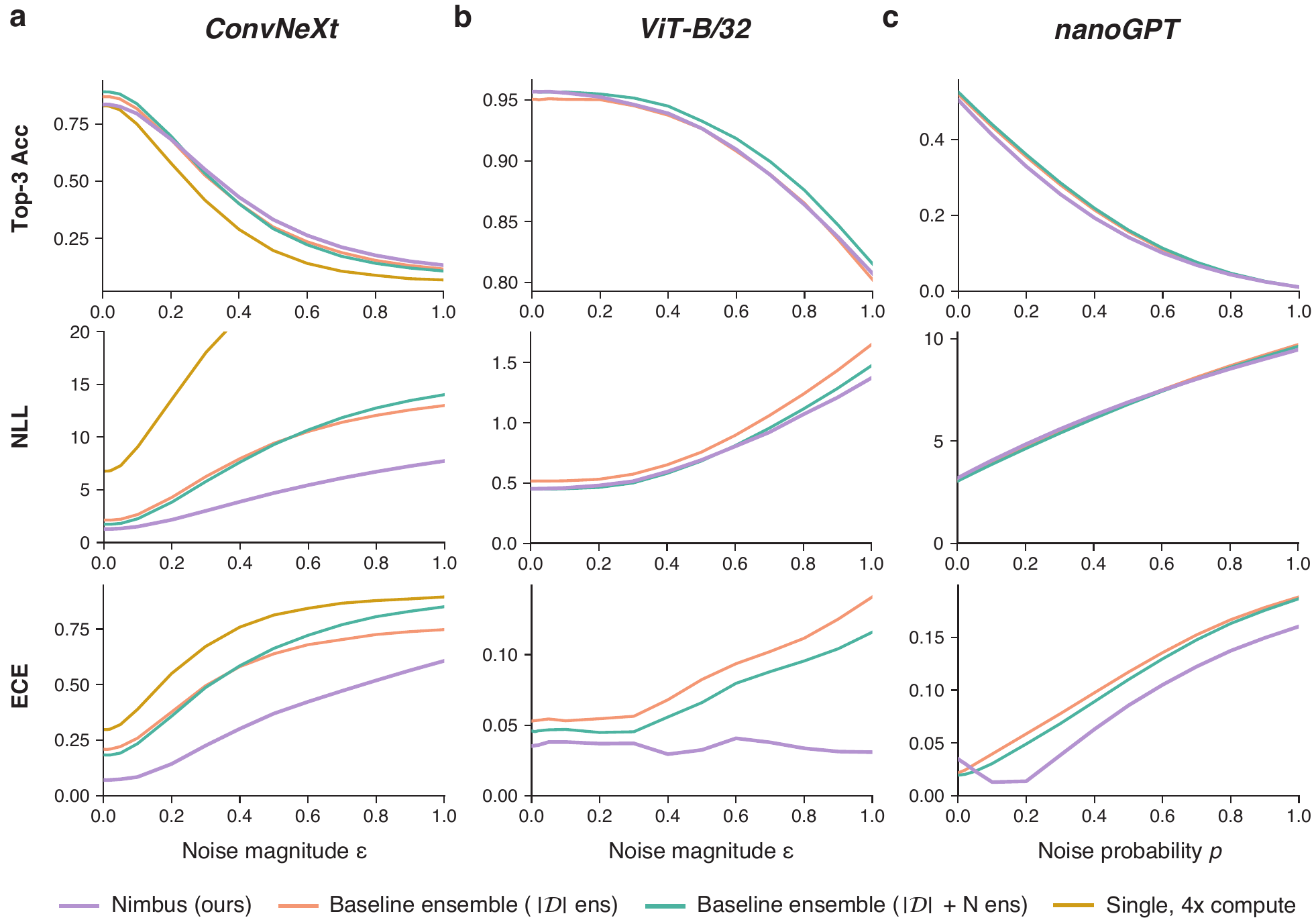}
    \caption{
    \textbf{\Nimbus is more robust to input perturbations than baselines.}
    \textbf{a.} ConvNeXt ($N = 7$), \textbf{b.} ViT-B/32 ($N = 8$), and \textbf{c.} nanoGPT ($N = 7$) robustness to input perturbations under logit ensembling for metrics such as top-3 accuracy (higher is better), NLL (lower is better), and ECE (lower is better). We compare between \Nimbus (purple) versus a baseline ensemble of $|\mathcal{D}|$ jointly trained models (orange) and $|\mathcal{D}| + N$ jointly trained models (green), and versus a single jointly trained model with $4 \times$ the effective compute (gold) as \Nimbus (only computationally feasible to run for ConvNeXt). \Nimbus finds a more \emph{functionally flat} manifold than baselines.
    }
    \label{fig:input_robustness}
\end{figure}

Relative to every compute-matched and extra-compute baseline, \Nimbus instantiated as ConvNeXt and ViT degrades more gracefully, with the largest gaps in NLL and ECE (see Fig.~\ref{fig:input_robustness}a--b). On language modeling, \Nimbus is comparable to or better than the baselines while achieving superior calibration (see Fig.~\ref{fig:input_robustness}c). \Nimbus thus produces a more \emph{functionally flat} manifold than baselines, crucially without any explicit optimization for functional flatness. 

\begin{figure}[h!]
\centering
    \includegraphics[width=0.80\textwidth]{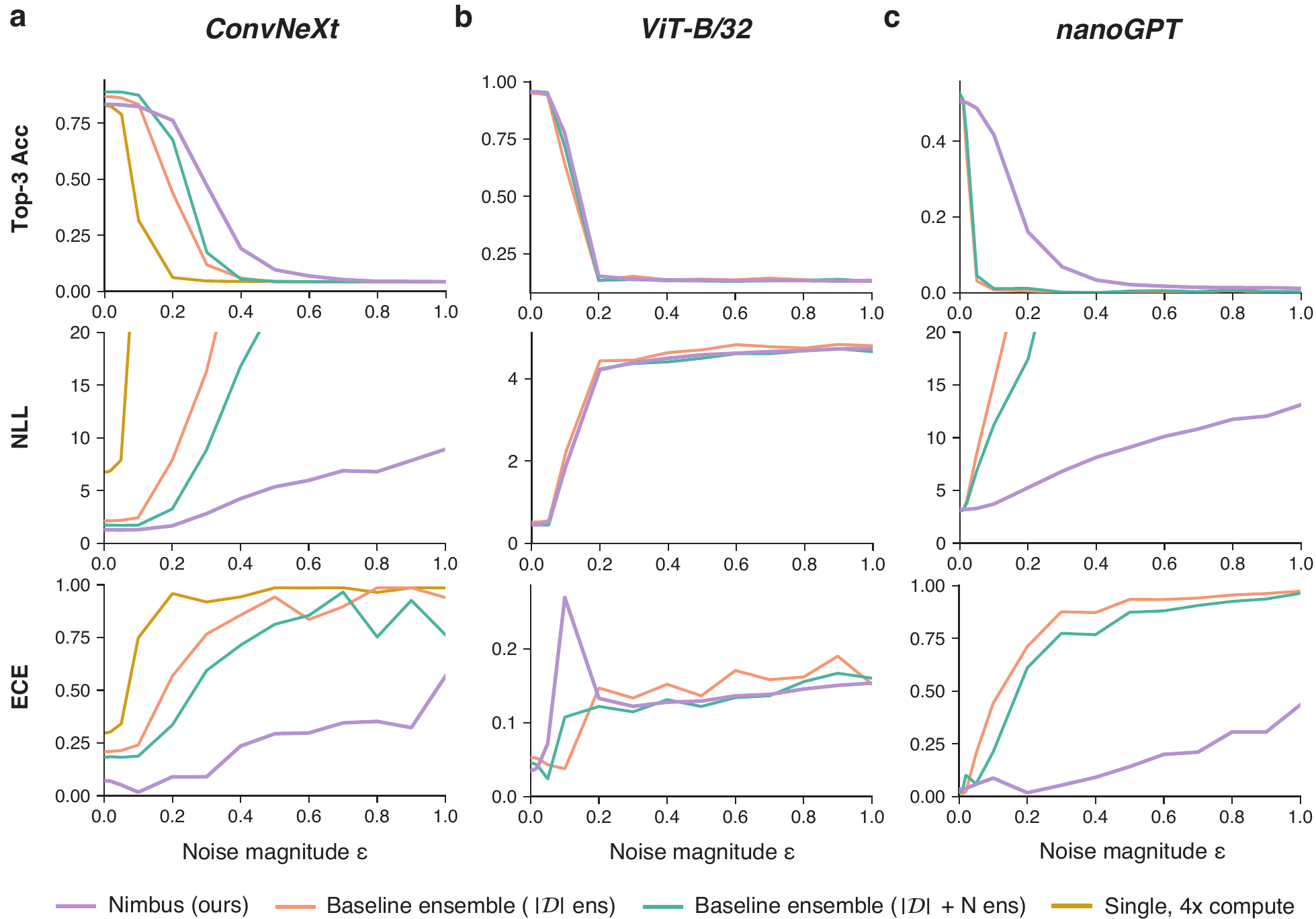}
    \caption{
    \textbf{\Nimbus is more robust to weight perturbations than baselines.}
    \textbf{a.} ConvNeXt ($N = 7$), \textbf{b.} ViT-B/32 ($N = 8$), and \textbf{c.} nanoGPT ($N = 7$) robustness to weight perturbations under logit ensembling for metrics such as top-3 accuracy (higher is better), NLL (lower is better), and ECE (lower is better). 
    Weight perturbations are additive Gaussian noise $\varepsilon$ relative to the norm of the model's own weights.  
    We compare between \Nimbus (purple) versus a baseline ensemble of $|\mathcal{D}|$ jointly trained models (orange) and $|\mathcal{D}| + N$ jointly trained models (green), and versus a single jointly trained model with $4 \times$ the effective compute (gold) as \Nimbus (only computationally feasible to run for ConvNeXt). \Nimbus finds a more \emph{weight-flat} manifold than baselines.
    }
    \label{fig:weight_robustness}
\end{figure}

\paragraph{Weight perturbations.}
Across all modalities, we perturb model weights with additive Gaussian noise of magnitude $\varepsilon$ relative to the $\ell_2$-norm of the full model. 
We again compare \Nimbus against every compute-matched and extra-compute baseline, and find that on ConvNeXt and nanoGPT, \Nimbus degrades significantly more gracefully, with the largest gaps in NLL and ECE (see Fig.~\ref{fig:weight_robustness}a, c). \Nimbus even outperforms ASAM, which is explicitly optimized for finding weight-flat minima (see Supp. Sec.~\ref{sec:full_experiments}). 
On ViT-B/32, \Nimbus is comparable to or better than the baselines for varying levels of weight noise (see Fig.~\ref{fig:weight_robustness}b).

We include a sensitivity analysis of our input and weight robustness experiments in Supp. Sec.~\ref{sec:sensitivity}. 

\section{Discussion and future work}

\paragraph{\Nimbus as a manifold learning framework for merging and ensembling.}
Where prior connectivity work links specialists by one-dimensional paths or two-dimensional sheets between a handful of same-objective networks, \Nimbus shows that an $(N-1)$-dimensional manifold of uniformly low loss exists between $N$ \emph{disjointly} trained ones, effectively reframing model merging as a traversable geometry instead of producing a single solution. As a consequence, the explicit manifolds that \Nimbus builds are also superior for robustness through arbitrarily large ensembles. 
As we show, our method naturally extends to post-training of transformer models. Since fine-tuned specialists share a pre-training recipe, the linear sum $\sum_i w_i \theta_i$ naturally lies close to a competent initialization, so the correction terms need only repair residual interference rather than resolve systemic neuron asymmetries.
A potential future direction is utilizing only the KL term in Algorithm~\ref{alg:nimbus}'s objective; eliminating the cross-entropy weight would test how much of the manifold's behavior can be carried through distillation alone, extending \Nimbus to regimes where source labels are scarce or unavailable.

\paragraph{A neural reading of the framework.}
The vertices $\{\theta_i\}$ map naturally onto specialized neural pathways. 
Biology realizes flexible degeneracy through localized mechanisms such as dendritic compartmentalization~\citep{otor2022dynamic, magee2020synaptic}, synaptic clustering~\citep{weber2016location, hedrick2022learning, hedrick2024local}, and the activation of silent synapses~\citep{vardalaki2025silent}, all of which adapt circuit computation without overwriting existing pathways.
\Nimbus approaches this problem through a deep learning lens.
Just as specialized circuits undergo structural changes to accumulate new capabilities, specialist models at the vertices of \Nimbus's manifold move away from boundaries to smoothly accumulate new task competencies.
Since a continuous family of models lies on the high-dimensional manifold induced by \Nimbus, a small synaptic perturbation can reach equally competent configurations with a minimal metabolic cost of synaptic change.
Our work fundamentally unifies concepts from neural degeneracy with mode connectivity and model merging, yielding a deep learning framework for flexible degeneracy.
This flexible degeneracy appears in both vision and language tasks, as \Nimbus is both functionally and locally weight-flat, with nearby points on the high-dimensional manifold close in parameter space and matched in behavior.

\paragraph{Limitations.}
While \Nimbus is quite performant in a within-modality heterogeneous setting (ViT fine-tuning tasks are visually unrelated domains), we have yet to test \Nimbus in a truly cross-modal setting, such as across language and vision simultaneously. 
Additionally, we are still searching for a core theoretical explanation of why or when constructing such a \Nimbus manifold is possible among arbitrary models.

\section*{Acknowledgements}

AI would like to thank Cynthia Rais, Raul Mojica Soto-Albors, Courtney Yaeger, and Akhilan Boopathy for invaluable help in figure making and for very useful scientific discussions. AI was supported by the MathWorks Graduate Research Fellowship. 
SC acknowledges support of the Department of Atomic Energy, Government of India, under Project No. RTI4001, the Murty Science Fellowship, and the Centre for Artificial Learning and Intelligence for Biological Research and Education, ICTS-TIFR. 

\bibliographystyle{unsrtnat}
\bibliography{bibliography}

@inproceedings{kobayashi2019simplex,
  title={B{\'e}zier simplex fitting: Describing Pareto fronts of simplicial problems with small samples in multi-objective optimization},
  author={Kobayashi, Ken and Hamada, Naoki and Sannai, Akiyoshi and Tanaka, Akinori and Bannai, Kenichi and Sugiyama, Masashi},
  booktitle={Proceedings of the AAAI Conference on Artificial Intelligence},
  volume={33},
  number={01},
  pages={2304--2313},
  year={2019}
}

@article{weber2016location,
  title={Location-dependent synaptic plasticity rules by dendritic spine cooperativity},
  author={Weber, Jens P and Andr{\'a}sfalvy, Bertalan K and Polito, Marina and Mag{\'o}, {\'A}d{\'a}m and Ujfalussy, Bal{\'a}zs B and Makara, Judit K},
  journal={Nature communications},
  volume={7},
  number={1},
  pages={11380},
  year={2016},
  publisher={Nature Publishing Group UK London}
}

@article{vardalaki2025silent,
  title={Silent synapses in the adult brain},
  author={Vardalaki, Dimitra and Yaeger, Courtney E and Harnett, Mark T},
  journal={Annual Review of Neuroscience},
  volume={48},
  year={2025},
  publisher={Annual Reviews}
}

@article{tononi1999measures,
  title={Measures of degeneracy and redundancy in biological networks},
  author={Tononi, Giulio and Sporns, Olaf and Edelman, Gerald M},
  journal={Proceedings of the National Academy of Sciences},
  volume={96},
  number={6},
  pages={3257--3262},
  year={1999},
  publisher={The National Academy of Sciences}
}

@article{magee2020synaptic,
  title={Synaptic plasticity forms and functions},
  author={Magee, Jeffrey C and Grienberger, Christine},
  journal={Annual review of neuroscience},
  volume={43},
  number={1},
  pages={95--117},
  year={2020},
  publisher={Annual Reviews}
}

@article{otor2022dynamic,
  title={Dynamic compartmental computations in tuft dendrites of layer 5 neurons during motor behavior},
  author={Otor, Yara and Achvat, Shay and Cermak, Nathan and Benisty, Hadas and Abboud, Maisan and Barak, Omri and Schiller, Yitzhak and Poleg-Polsky, Alon and Schiller, Jackie},
  journal={Science},
  volume={376},
  number={6590},
  pages={267--275},
  year={2022},
  publisher={American Association for the Advancement of Science}
}

@article{hedrick2022learning,
  title={Learning binds new inputs into functional synaptic clusters via spinogenesis},
  author={Hedrick, Nathan G and Lu, Zhongmin and Bushong, Eric and Singhi, Surbhi and Nguyen, Peter and Maga{\~n}a, Yessenia and Jilani, Sayyed and Lim, Byung Kook and Ellisman, Mark and Komiyama, Takaki},
  journal={Nature neuroscience},
  volume={25},
  number={6},
  pages={726--737},
  year={2022},
  publisher={Nature Publishing Group US New York}
}

@article{hedrick2024local,
  title={Local and global predictors of synapse elimination during motor learning},
  author={Hedrick, Nathan G and Wright, William J and Komiyama, Takaki},
  journal={Science Advances},
  volume={10},
  number={11},
  pages={eadk0540},
  year={2024},
  publisher={American Association for the Advancement of Science}
}

@misc{Karpathy2022,
  author = {Andrej Karpathy},
  title = {\text{NanoGPT}},
  year = {2022},
  publisher = {GitHub},
  journal = {GitHub repository},
  howpublished = {\url{https://github.com/karpathy/nanoGPT}},
  commit = {325be85d9be8c81b436728a420e85796c57dba7e}
}

@inproceedings{liu2022convnet,
  title={A convnet for the 2020s},
  author={Liu, Zhuang and Mao, Hanzi and Wu, Chao-Yuan and Feichtenhofer, Christoph and Darrell, Trevor and Xie, Saining},
  booktitle={Proceedings of the IEEE/CVF conference on computer vision and pattern recognition},
  pages={11976--11986},
  year={2022}
}

@article{tinystories,
  title={Tinystories: How small can language models be and still speak coherent english?},
  author={Eldan, Ronen and Li, Yuanzhi},
  journal={arXiv preprint arXiv:2305.07759},
  year={2023}
}

@article{wikitext,
  title={Pointer sentinel mixture models},
  author={Merity, Stephen and Xiong, Caiming and Bradbury, James and Socher, Richard},
  journal={arXiv preprint arXiv:1609.07843},
  year={2016}
}

@misc{gutenberg_website,
  author = {{Project Gutenberg}},
  title = {Project {G}utenberg},
  year = {2026},
  howpublished = {\url{https://www.gutenberg.org}},
  note = {Accessed: May 2, 2026}
}

@software{cpython,
  author       = {{Python Software Foundation}},
  title        = {CPython},
  url          = {https://github.com/python/cpython},
  year         = {2026},
  note         = {Accessed: 2026-05-02}
}

@inproceedings{henderson2022pileoflaw,
    title={Pile of Law: Learning Responsible Data Filtering from the Law and a 256{GB} Open-Source Legal Dataset},
    author={Henderson, Peter and Krass, Mark S. and Zheng, Lucia and Guha, Neel and Manning, Christopher D. and Jurafsky, Dan and Ho, Daniel E.},
    booktitle={Advances in Neural Information Processing Systems (NeurIPS)},
    volume={35},
    pages={29217--29234},
    year={2022}
}

@article{he2019arxiv,
    title={Long Document Classification From Local Word Glimpses via Recurrent Attention Learning},
    author={He, Jun and Wang, Liqun and Liu, Liu and Feng, Jiao and Wu, Hao},
    journal={IEEE Access},
    volume={7},
    pages={40707--40718},
    year={2019},
    publisher={IEEE},
    doi={10.1109/ACCESS.2019.2907992}
}

@misc{codeparrot,
    author       = {{CodeParrot (Hugging Face)}},
    title        = {{CodeParrot}: A Cleaned {P}ython Code Dataset},
    year         = {2022},
    howpublished = {\url{https://huggingface.co/datasets/codeparrot/codeparrot-clean}},
    note         = {Accessed: 2026-05-02}
}

@inproceedings{frankle2020linear,
  title={Linear mode connectivity and the lottery ticket hypothesis},
  author={Frankle, Jonathan and Dziugaite, Gintare Karolina and Roy, Daniel and Carbin, Michael},
  booktitle={International conference on machine learning},
  pages={3259--3269},
  year={2020},
  organization={PMLR}
}

@inproceedings{wortsman2022model,
  title={Model soups: averaging weights of multiple fine-tuned models improves accuracy without increasing inference time},
  author={Wortsman, Mitchell and Ilharco, Gabriel and Gadre, Samir Ya and Roelofs, Rebecca and Gontijo-Lopes, Raphael and Morcos, Ari S and Namkoong, Hongseok and Farhadi, Ali and Carmon, Yair and Kornblith, Simon and others},
  booktitle={International conference on machine learning},
  pages={23965--23998},
  year={2022},
  organization={PMLR}
}

@inproceedings{mode_connectivity_bezier,
  title={Revisiting Mode Connectivity in Neural Networks with Bezier Surface.},
  author={Ren, Jie and Chen, Pin-Yu and Wang, Ren},
  booktitle={ICLR},
  year={2025}
}

@inproceedings{nguyen2019connected,
  title={On connected sublevel sets in deep learning},
  author={Nguyen, Quynh},
  booktitle={International conference on machine learning},
  pages={4790--4799},
  year={2019},
  organization={PMLR}
}

@inproceedings{asam,
  title={Asam: Adaptive sharpness-aware minimization for scale-invariant learning of deep neural networks},
  author={Kwon, Jungmin and Kim, Jeongseop and Park, Hyunseo and Choi, In Kwon},
  booktitle={International conference on machine learning},
  pages={5905--5914},
  year={2021},
  organization={PMLR}
}

@article{zhou2023going,
  title={Going beyond linear mode connectivity: The layerwise linear feature connectivity},
  author={Zhou, Zhanpeng and Yang, Yongyi and Yang, Xiaojiang and Yan, Junchi and Hu, Wei},
  journal={Advances in neural information processing systems},
  volume={36},
  pages={60853--60877},
  year={2023}
}

@article{garipov2018loss,
  title={Loss surfaces, mode connectivity, and fast ensembling of dnns},
  author={Garipov, Timur and Izmailov, Pavel and Podoprikhin, Dmitrii and Vetrov, Dmitry P and Wilson, Andrew G},
  journal={Advances in neural information processing systems},
  volume={31},
  year={2018}
}

@article{gitrebasin,
  title={Git re-basin: Merging models modulo permutation symmetries},
  author={Ainsworth, Samuel K and Hayase, Jonathan and Srinivasa, Siddhartha},
  journal={arXiv preprint arXiv:2209.04836},
  year={2022}
}

@article{swa,
  title={Averaging weights leads to wider optima and better generalization},
  author={Izmailov, Pavel and Podoprikhin, Dmitrii and Garipov, Timur and Vetrov, Dmitry and Wilson, Andrew Gordon},
  journal={arXiv preprint arXiv:1803.05407},
  year={2018}
}

@inproceedings{benton2021loss,
  title={Loss surface simplexes for mode connecting volumes and fast ensembling},
  author={Benton, Gregory and Maddox, Wesley and Lotfi, Sanae and Wilson, Andrew Gordon Gordon},
  booktitle={International Conference on Machine Learning},
  pages={769--779},
  year={2021},
  organization={PMLR}
}

@article{fisher_merging,
  title={Merging models with fisher-weighted averaging},
  author={Matena, Michael S and Raffel, Colin A},
  journal={Advances in Neural Information Processing Systems},
  volume={35},
  pages={17703--17716},
  year={2022}
}

@article{ovadia2019can,
  title={Can you trust your model's uncertainty? evaluating predictive uncertainty under dataset shift},
  author={Ovadia, Yaniv and Fertig, Emily and Ren, Jie and Nado, Zachary and Sculley, David and Nowozin, Sebastian and Dillon, Joshua and Lakshminarayanan, Balaji and Snoek, Jasper},
  journal={Advances in neural information processing systems},
  volume={32},
  year={2019}
}

@article{zipit,
  title={Zipit! merging models from different tasks without training},
  author={Stoica, George and Bolya, Daniel and Bjorner, Jakob and Ramesh, Pratik and Hearn, Taylor and Hoffman, Judy},
  journal={arXiv preprint arXiv:2305.03053},
  year={2023}
}

@article{ties_merging,
  title={Ties-merging: Resolving interference when merging models},
  author={Yadav, Prateek and Tam, Derek and Choshen, Leshem and Raffel, Colin A and Bansal, Mohit},
  journal={Advances in neural information processing systems},
  volume={36},
  pages={7093--7115},
  year={2023}
}

@article{pcgrad,
  title={Gradient surgery for multi-task learning},
  author={Yu, Tianhe and Kumar, Saurabh and Gupta, Abhishek and Levine, Sergey and Hausman, Karol and Finn, Chelsea},
  journal={Advances in neural information processing systems},
  volume={33},
  pages={5824--5836},
  year={2020}
}

@inproceedings{draxler2018essentially,
  title={Essentially no barriers in neural network energy landscape},
  author={Draxler, Felix and Veschgini, Kambis and Salmhofer, Manfred and Hamprecht, Fred},
  booktitle={International conference on machine learning},
  pages={1309--1318},
  year={2018},
  organization={PMLR}
}

@article{ilharco2022editing,
  title={Editing models with task arithmetic},
  author={Ilharco, Gabriel and Ribeiro, Marco Tulio and Wortsman, Mitchell and Gururangan, Suchin and Schmidt, Ludwig and Hajishirzi, Hannaneh and Farhadi, Ali},
  journal={arXiv preprint arXiv:2212.04089},
  year={2022}
}

@article{jin2022dataless,
  title={Dataless knowledge fusion by merging weights of language models},
  author={Jin, Xisen and Ren, Xiang and Preotiuc-Pietro, Daniel and Cheng, Pengxiang},
  journal={arXiv preprint arXiv:2212.09849},
  year={2022}
}

@inproceedings{yu2024language,
  title={Language models are super mario: Absorbing abilities from homologous models as a free lunch},
  author={Yu, Le and Yu, Bowen and Yu, Haiyang and Huang, Fei and Li, Yongbin},
  booktitle={Forty-first International Conference on Machine Learning},
  year={2024}
}

@article{yang2023adamerging,
  title={Adamerging: Adaptive model merging for multi-task learning},
  author={Yang, Enneng and Wang, Zhenyi and Shen, Li and Liu, Shiwei and Guo, Guibing and Wang, Xingwei and Tao, Dacheng},
  journal={arXiv preprint arXiv:2310.02575},
  year={2023}
}

@article{lakshminarayanan2017simple,
  title={Simple and scalable predictive uncertainty estimation using deep ensembles},
  author={Lakshminarayanan, Balaji and Pritzel, Alexander and Blundell, Charles},
  journal={Advances in neural information processing systems},
  volume={30},
  year={2017}
}

@inproceedings{guo2017calibration,
  title={On calibration of modern neural networks},
  author={Guo, Chuan and Pleiss, Geoff and Sun, Yu and Weinberger, Kilian Q},
  booktitle={International conference on machine learning},
  pages={1321--1330},
  year={2017},
  organization={PMLR}
}

@InProceedings{radford2021clip,
  title = 	 {Learning Transferable Visual Models From Natural Language Supervision},
  author =       {Radford, Alec and Kim, Jong Wook and Hallacy, Chris and Ramesh, Aditya and Goh, Gabriel and Agarwal, Sandhini and Sastry, Girish and Askell, Amanda and Mishkin, Pamela and Clark, Jack and Krueger, Gretchen and Sutskever, Ilya},
  booktitle = 	 {Proceedings of the 38th International Conference on Machine Learning},
  pages = 	 {8748--8763},
  year = 	 {2021},
  editor = 	 {Meila, Marina and Zhang, Tong},
  volume = 	 {139},
  series = 	 {Proceedings of Machine Learning Research},
  month = 	 {18--24 Jul},
  publisher =    {PMLR},
  url = 	 {https://proceedings.mlr.press/v139/radford21a.html}
}

@article{cifar100,
  title={Learning multiple layers of features from tiny images},
  author={Krizhevsky, Alex and Hinton, Geoffrey and others},
  year={2009},
  publisher={Toronto, ON, Canada}
}

@article{lora,
  title={Lora: Low-rank adaptation of large language models},
  author={Hu, Edward J and Shen, Yelong and Wallis, Phillip and Allen-Zhu, Zeyuan and Li, Yuanzhi and Wang, Shean and Wang, Lu and Chen, Weizhu},
  journal={arXiv preprint arXiv:2106.09685},
  year={2021}
}

\newpage
\appendix

\section{Experimental details}

\subsection{Hyperparameters}
\label{sec:hyperparams}

We include all the relevant hyperparameters from our experiments in Table~\ref{tab:hyperparams}. Our code can be found at \texttt{https://github.com/abhi-iyer/nimbus}. All experiments were run on a single NVIDIA RTX 4090 GPU; in total, all experiments took about 10--12 hours to run.

\begingroup
\small
\setlength{\tabcolsep}{4pt}
\renewcommand{\arraystretch}{1.05}

\begin{xltabular}{\linewidth}{@{}l >{\raggedleft\arraybackslash}X >{\raggedleft\arraybackslash}X >{\raggedleft\arraybackslash}X@{}}
\toprule
 & \textbf{CIFAR-100} & \textbf{ViT} & \textbf{Language modeling} \\
\midrule
\endfirsthead
\multicolumn{4}{@{}l}{\emph{Table~\ref{tab:hyperparams} (continued)}}\\
\toprule
 & \textbf{CIFAR-100} & \textbf{ViT} & \textbf{Language modeling} \\
\midrule
\endhead
\midrule
\multicolumn{4}{r@{}}{\footnotesize\emph{continued on next page}}\\
\endfoot
\bottomrule
\caption{\Nimbus hyperparameters, parameters, and compute across all three experiments.}
\label{tab:hyperparams}\\
\endlastfoot
\multicolumn{4}{@{}l}{\emph{Architecture}} \\
\quad Backbone            & SmallConvNeXt      & CLIP ViT-B/32 & nanoGPT \\
\quad Layers / depth      & depths $(1,1,3)$   & $12$ blocks & $n_\mathrm{layer}=4$ \\
\quad Width               & dims $(16,48,192)$ & $768$ & $n_\mathrm{embed}=128$ \\
\quad Heads / expansion   & expand $(2,2,3)$   & $12$ heads & $n_\mathrm{head}=4$ \\
\quad MLP                 & ---                & $3072$ & $4\times$ \\
\quad Patch / block size  & ---                & patch $32$, image $224$ & block $256$ \\
\quad Activation          & SiLU               & QuickGELU & GELU \\
\quad Dropout             & ---                & $0$ & $0.1$ \\
\quad Vocab / classes     & $10N$              & zero-shot text heads ($512$-dim output) & $2048$ \\
\quad Tokenizer           & ---                & --- & byte-level BPE \\
\midrule
\multicolumn{4}{@{}l}{\emph{Vertex training (given as input to Algorithm~\ref{alg:nimbus})}} \\
\quad Iters / epochs      & $100$ epochs        & $16$k steps (early-stop) & $100$k iters (early-stop) \\
\quad Batch (per task)    & $256$               & $64$ & $64$ \\
\quad Optimizer           & AdamW $(0.9,0.999)$ & AdamW $(0.9,0.95)$ & AdamW $(0.9,0.95)$ \\
\quad Weight decay        & $0$                 & $0$ & $0$ \\
\quad LR / schedule       & 8e-3, OneCycleLR (pct\_start $0.2$) & 1e-4, warmup $100\to$ const & 6e-4, warmup $200\to$ const \\
\quad Vertex form         & full model          & rank-$64$ LoRA (on frozen CLIP) & full model \\
\midrule
\multicolumn{4}{@{}l}{\emph{Phase 1 -- training correction tensors}} \\
\quad Iters / epochs (per $r\in\mathcal{D}$) & $100$ epochs & $1000$ iters & $6000$ iters \\
\quad Batch (per task)    & $256$               & $64$ & $64$ \\
\quad Optimizer           & AdamW $(0.9,0.999)$ & AdamW $(0.9,0.95)$ & AdamW $(0.9,0.95)$ \\
\quad LR $\eta_1$ (constant) & 8e-3             & 5e-5 & 8e-3 \\
\quad Weight decay        & $0$                 & $0$ & $0$ \\
\quad Correction tensor form           & dense               & rank-$64$ LoRA $+$ dense & dense \\
\midrule
\multicolumn{4}{@{}l}{\emph{Phase 2 -- gate calibration}} \\
\quad Iters / epochs      & $50$ epochs         & $300$ iters & $1000$ iters \\
\quad Batch               & $256$               & $64$ & $64$ \\
\quad Optimizer           & AdamW $(0.9,0.999)$ & AdamW $(0.9,0.95)$ & AdamW $(0.9,0.95)$ \\
\quad LR $\eta_2$ (constant) & 1e-3             & 2e-3 & 2e-3 \\
\quad Samples / step $S$  & $5$                 & $4$ & $4$ \\
\midrule
\multicolumn{4}{@{}l}{\emph{Evaluation}} \\
\quad Total $\bm{w}$ samples & $100$            & $100$ & $100$ \\
\midrule
\multicolumn{4}{@{}l}{\emph{Trainable parameters (M)}} \\
\quad \Nimbus                       & $1.46/2.19/2.21/2.22$ & $39.4$ ($126.9$ total)  & $3.25$ \\
\quad Baseline ensemble ($|\mathcal{D}|$)     & $1.46/2.19/2.21/2.22$ & $39.4$ ($126.9$ total)  & $3.25$ \\
\quad Baseline ensemble ($|\mathcal{D}|{+}N$) & $3.64/5.85/7.35/9.63$ & $144.5$ ($232.0$ total) & $10.8$ \\
\quad Single, 4$\times$ compute    & $2.88/4.32/4.33/4.34$ & ---                     & --- \\
\midrule
\multicolumn{4}{@{}l}{\emph{Phase-1 sample-passes (M)}} \\
\quad \Nimbus                       & $12.2/47.3/85.1/162$  & $15.36$ & $72.6$ \\
\quad Baseline ensemble ($|\mathcal{D}|$)     & $12.2/47.3/85.1/162$  & $15.36$ & $72.6$ \\
\quad Baseline ensemble ($|\mathcal{D}|{+}N$) & $30.4/126/284/702$    & $56.3$  & $241.9$ \\
\quad Single, 4$\times$ compute    & $24.3/94.5/170.1/324$ & ---     & --- \\
\midrule
\multicolumn{4}{@{}l}{\emph{Tasks and data}} \\
\quad Source              & CIFAR-100 & 8 tasks from Task Arithmetic~\citep{ilharco2022editing} & 7 text domains \\
\quad $(N,\,\mathcal{D})$ pairs
  & \makecell[tr]{$N{=}3$:\ $\{2,3\}$ \\ $N{=}5$:\ $\{2,4,5\}$ \\
                  $N{=}7$:\ $\{2,6,7\}$ \\ $N{=}10$:\ $\{2,9,10\}$}
  & $N{=}8$:\ $\{2,7,8\}$ & $N{=}7$:\ $\{2,6,7\}$ \\
\quad Per-task budget     & $10$ classes & $\le 50$k images & $20$M tokens ($60$M chars) \\
\quad Augmentation        & reflect-pad $4$, random $32$ crop, h-flip ($p{=}0.5$) & CLIP preprocess (resize $224$, normalize) & none \\
\quad LoRA rank $r$       & --- & $64$ & --- \\
\quad Random seed         & $42$ & $42$ & $42$ \\
\end{xltabular}
\endgroup

\subsection{Compute-matched baselines}
\label{sec:compute_matching}

As a first-order approximation, \Nimbus requires:

\vspace{-20pt}
\begin{equation*}
    \text{(passes per training step)} \cdot \text{(epochs or iters / degree)} \cdot \text{(batch size)} \sum_{r \in \mathcal{D}} \min \left( \binom{N}{r}, N \right) \cdot r
\end{equation*}
\vspace{-22pt}

``sample-passes'' (training samples $\times$ passes). Following standard FLOP counting estimates, we define the ``passes per training step'' as $3$ (wherein we approximate each forward through a network as $1$ ``pass'' and backward as $2$ ``passes''). The binomial term measures the number of faces per degree $r$ that we wish to sample from, while the second $r$ term measures the passes per task per face (i.e., a degree-$r$ face joins $r$ task-specialized models). 

For example, if our problem setting is ViT-B/32 with $N = 8$, $\mathcal{D} = \{2, 7, 8\}$, $\text{iters / degree} = 1000$, and $\text{batch size} = 64$, then this reduces to $\approx 15.36$M sample-passes. We only include the sample-passes consumed by the degree-by-degree training phase, as the calibration phase is relatively much cheaper. Additionally, \Nimbus with ViT-B/32 yields approximately $39.4$M learnable parameters. Since $|\mathcal{D}| = 3$, an equitable baseline fine-tunes $3$ different pre-trained ViT-B/32 base models jointly on all the tasks and ensembles over them. The LoRA rank is likewise $64$ (total learnable parameters across baseline models are equal to the total parameter budget of \Nimbus, which is $39.4$M). The number of iterations is picked such that the total number of sample-passes across baseline models is $15.36$M (each baseline model receives $5.12$M sample-passes). We call this compute-matched baseline the ``baseline ensemble'' with $|\mathcal{D}|$ member models. 

If we suppose that a practitioner employing \Nimbus must also assume responsibility for the compute in training the vertices, we also include a ``baseline ensemble'' with $|\mathcal{D}| + N$ member models (each trained with the same number of sample-passes as the original ``baseline ensemble'') to liberally account for this extra compute (effectively $\frac{N}{|\mathcal{D}|} \times$ \emph{more} compute). And if we further suppose that a practitioner does not want to ensemble at all, we also include a single baseline model that is jointly trained on all the tasks with $2 \times$ the sample-passes and $2 \times$ the total learnable parameters of \Nimbus (effectively $4 \times$ the compute).

\section{Pseudocode}
\label{sec:pseudocode}

\begin{algorithm}[H]
\caption{\textbf{\Nimbus}: \underline{N}-expert \underline{I}nterpolating \underline{M}anifold: \underline{B}ezier over an \underline{U}nderlying \underline{S}implex}
\label{alg:nimbus}
\begin{algorithmic}[1]
\Require Vertex models $\{\theta_i\}_{i=1}^{N}$, datasets $\{\mathcal{T}_i\}$, degree set $\mathcal{D} \subseteq \{2,\dots,N\}$, learning rates $\eta_1, \eta_2$, iteration budgets $T_1, T_2$, calibration sample count $S$.
\State Initialize $\tilde{P}_r \gets \bm{0}$ and $g_r \gets 0$ for all $r \in \mathcal{D}$.
\Statex
\Statex \hspace{-1.2em}\textbf{Phase 1: low-to-high degree hierarchical training}
\For{$r \in \mathcal{D}$ in increasing order}
    \State Make $\tilde{P}_r$ and $\{g_{r'} : r' < r,\, r' \in \mathcal{D}\}$ trainable; freeze the rest ($g_r$ stays at $0$).
    \For{$t = 1, \dots, T_1$}
        \State Draw $(x_i, y_i) \sim \mathcal{T}_i$ for each $i = 1, \dots, N$.
        \State Sample $\min(N, \binom{N}{r})$ multi-indices $\{\bm{\alpha}\}$ of degree $r$ uniformly without replacement.
        \For{each sampled $\bm{\alpha}$ with support $F = \{i : \alpha_i = 1\}$}
            \State Sample $\bm{w}$ on face $F$;\; let $x_F, y_F = \mathrm{concat}_{i \in F}(x_i, y_i)$.
            \State $\mathcal{L}_{\bm{\alpha}} \gets \mathrm{CE}\bigl(f_{\theta(\bm{w})}(x_F),\, y_F\bigr) + \mathrm{KL}\!\bigl(\sigma(f_{\theta(\bm{w})}(x_F)) \,\big\|\, \sigma(\mathrm{concat}_{i \in F} f_{\theta_i}(x_i))\bigr)$.
        \EndFor
        \State One AdamW step at LR $\eta_1$ on $\mathbb{E}_{\bm{\alpha}}[\mathcal{L}_{\bm{\alpha}}]$.
    \EndFor
\EndFor
\Statex
\Statex \hspace{-1.2em}\textbf{Phase 2: gate calibration}
\State Freeze all $\tilde{P}_r$; make every gate $\{g_r\}_{r \in \mathcal{D}}$ trainable.
\For{$t = 1, \dots, T_2$}
    \State Draw $(x_i, y_i) \sim \mathcal{T}_i$ for each $i$;\; set $x, y = \mathrm{concat}_i(x_i, y_i)$.
    \State Sample $S$ multi-indices $\{\bm{\alpha}_s\}$ from $\bigcup_{r \in \mathcal{D}}\{\bm{\alpha} \in \{0,1\}^N : |\bm{\alpha}| = r\}$, equal weight per degree and uniform within degree.
    \State For each $s$: sample $\bm{w}_s$ on face $F_s$ and set $\ell_s \gets f_{\theta(\bm{w}_s)}(x)$.
    \State $\mathcal{L}_{\mathrm{NLL}} \gets \tfrac{1}{2}\bigl[\mathrm{CE}\bigl(\tfrac{1}{S}\!\sum_s \ell_s,\, y\bigr) - \tfrac{1}{|y|}\!\sum_j \log\bigl(\tfrac{1}{S}\!\sum_s \sigma(\ell_s)\bigr)_{\!j, y_j}\bigr]$.
    \For{$s = 1, \dots, S$}
        \State Let $\ell_s^{F_s}$ be the rows of $\ell_s$ for tasks in $F_s$.
        \State $\mathcal{K}_s \gets \mathrm{KL}\!\bigl(\sigma(\ell_s^{F_s}) \,\big\|\, \sigma(\mathrm{concat}_{i \in F_s} f_{\theta_i}(x_i))\bigr)$.
    \EndFor
    \State One AdamW step at LR $\eta_2$ on $\mathcal{L}_{\mathrm{NLL}} + \tfrac{1}{S}\!\sum_s \mathcal{K}_s$ (gates only).
\EndFor
\Statex
\State \Return $\{\tilde{P}_r,\, g_r\}_{r \in \mathcal{D}}$.
\end{algorithmic}
\end{algorithm}

\section{Ablations}

\subsection{Algorithm ablations}
\label{sec:algorithm_ablation}

\paragraph{Results.} Table~\ref{tab:algorithm_ablation} ablates the main design choices in \Nimbus for the ConvNeXt, $N=3$ setting. The full model achieves the lowest mean and worst-case NLL, indicating that both pairwise and highest-order corrections are needed to fill the simplex without leaving high-loss regions. Pairwise-only corrections ($\mathcal{D} = \{2\}$) substantially degrade mean NLL, suggesting that edge-level repairs alone are insufficient for the simplex interior. Conversely, the interior-only correction ($\mathcal{D} = \{3\}$) attains reasonable average NLL but has a very large worst-case NLL, consistent with the absence of learned structure on lower-dimensional faces. Surprisingly, separate per-face corrections (no sharing of correction tensors within a degree) perform worse than the shared per-degree parameterization despite being strictly more expressive. Additionally, Phase~2 crucially improves mean NLL while KL improves worst-case stability.

\paragraph{Why shared correction terms?}
Without shared correction terms (i.e., unique correction tensors for every face at every degree $d$), the number of learnable tensors would scale as $\mathcal{O}(d^{N-1})$ and become computationally prohibitive to optimize for any reasonable choice of $N$. 
Additionally, shared correction terms seem to act as a useful regularizer, as per-face corrections are a more ill-posed optimization problem within our hierarchical training recipe. 
Thus, the shared correction tensors across all faces of a particular degree emerged as a natural choice to operationalize \Nimbus. 

\paragraph{Why does the interaction order appear to matter more than the identities of the participating specialists?}
Because we are primarily interested in ensembling models along the $N$-th order \Bezier surface (the full surface constructed by all vertex models), we hypothesize that the main direction we wish to deform such a surface is described by the highest-order $N$-th term (which can be visualized as tugging the surface from its centroid). Working backwards, if we wish to add important error correction terms (corrective tensors whose order is strictly less than $N$), then the easiest way to do so is to add one corrective tensor per order, in which case such a correction is shared among all faces within an order. We suspect that the interaction order appears to matter more than the identities of the vertices because the \Bezier control tensors ($\tilde{P}_r$) are only intended to \emph{correct} the $N$-th order surface around the local neighborhood of the centroid.

Overall, these ablations support the full \Nimbus design: shared pairwise and highest-order corrections, stabilized by distillation and lightweight gate calibration, produce the most uniformly low-loss manifold. Robustness ablations follow the same trend across all variants as these no-perturbation ablations and are omitted for succinctness.

\begin{table}[H]
\centering
\small
\caption{
Ablation study for ConvNeXt, $N=3$. We report mean and worst-case NLL (lower is better) by sampling over the interior points of the simplex. The full \Nimbus model uses both pairwise and highest-order corrections, KL distillation, and Phase~2 gate calibration.
}
\label{tab:algorithm_ablation}
\begin{tabular}{lcc}
\toprule
\textbf{Variant} & \textbf{Mean NLL} & \textbf{Worst NLL} \\
\midrule
Full \Nimbus          & $\mathbf{1.3040}$ & $\mathbf{1.7548}$ \\
$\mathcal{D} = \{2\}$         & $1.8950$          & $2.6937$ \\
$\mathcal{D} = \{3\}$          & $1.7591$          & $6.2718$ \\
\Nimbus without degree-sharing     & $1.5013$          & $3.5682$ \\
No KL distillation    & $1.3680$          & $2.1651$ \\
No Phase~2            & $1.4803$          & $2.0609$ \\
\bottomrule
\end{tabular}
\end{table}

\subsection{Interaction order ablations}
\label{sec:interaction_ablation}

Another important aspect of scaling is to consider how many corrective tensors \Nimbus requires as we consider an increasing number of specialist models. We try to rigorously identify which corrections are needed for optimal performance of \Nimbus by ablating the choices for $\mathcal{D}$ for ConvNeXt, $N=7$.

\begin{figure}[H]
\centering
    \includegraphics[width=1.0\textwidth]{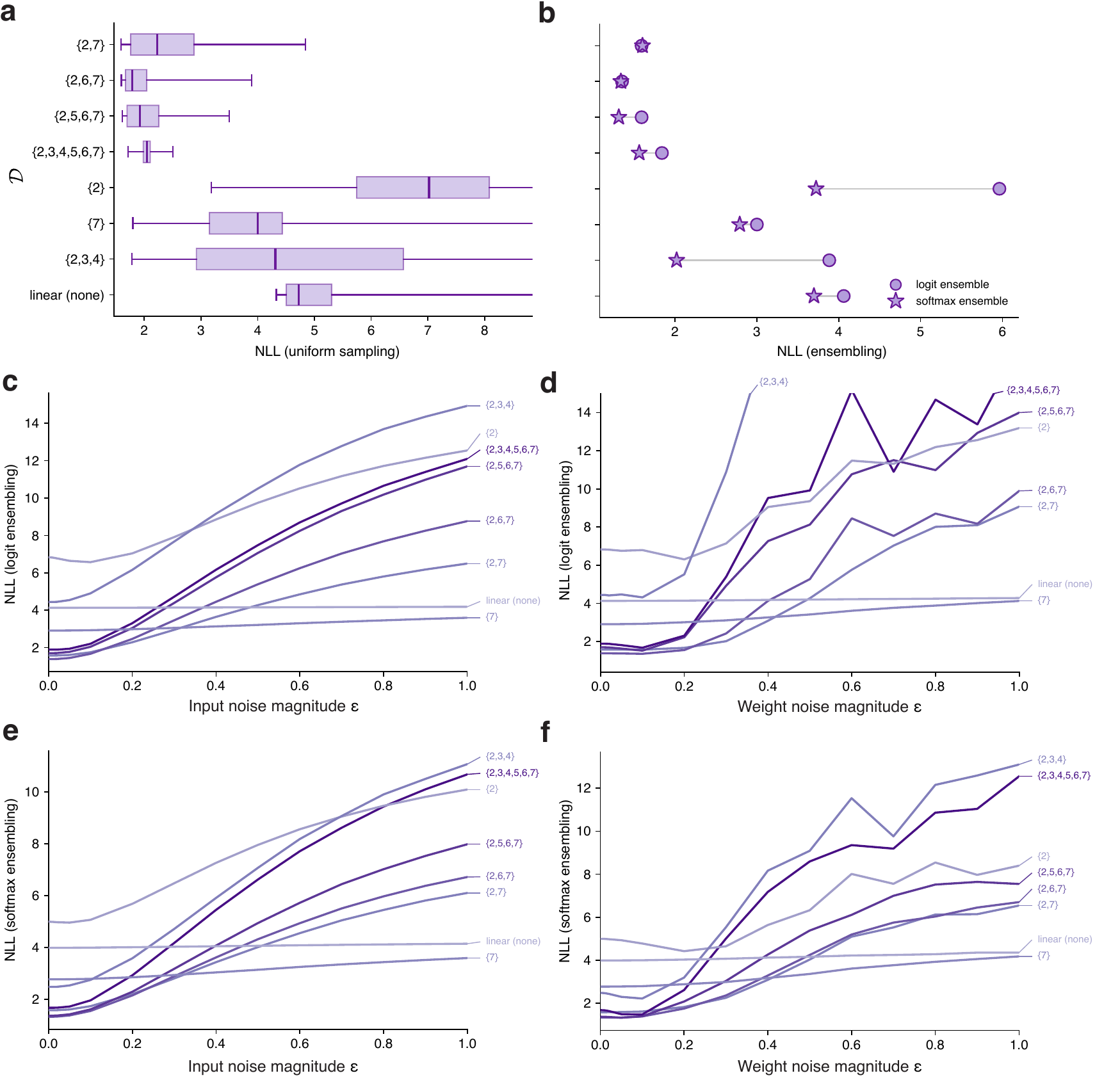}
    \caption{\textbf{Ablating the choice of interaction orders in \Nimbus.} \textbf{a.} Uniformly sampled multi-task NLL loss for various choices of $\mathcal{D}$. \textbf{b.} Under the same choices of $\mathcal{D}$, the gap between logit and softmax ensembling. \textbf{c, d.} Robustness to input and weight perturbations, respectively, under logit ensembling. \textbf{e, f.} Robustness to input and weight perturbations, respectively, under softmax ensembling.}
    \label{fig:supp_d_sweep}
\end{figure}

We uniformly sample over the constructed manifold for various choices of degrees (possible choices being subsets of $\{2, ..., 7\}$) and determine the multi-task loss (see Supp. Fig.~\ref{fig:supp_d_sweep}a). We also investigate how the multi-task NLL appears under logit vs. softmax ensembling for the same choices of degrees (see Supp. Fig.~\ref{fig:supp_d_sweep}b).

We see that $\mathcal{D} = \{2, 6, 7\}$ is clearly the optimal choice in terms of both the quality of the NLL loss distribution (tight spread around the mean) and space complexity (number of learnable parameters that scale with $|\mathcal{D}|$). Additionally, the gap between logit and softmax ensembling reduces significantly under our core hypothesis (lower degrees coupled with a few higher-order degrees are sufficient). 

We also showcase \Nimbus's gentle NLL loss degradation under both logit and softmax ensembling for each choice of $\mathcal{D}$ (see Supp. Fig.~\ref{fig:supp_d_sweep}c--f). $\mathcal{D} = \{2, 6, 7\}$ is optimal across all robustness curves, as it (1) starts off with the lowest NLL loss, (2) is the most robust against small-magnitude noise, and (3) degrades more gracefully with higher-magnitude noise at a fraction of the total learnable parameters (relative to $\{2, 5, 6, 7\}$ or $\{2, 3, 4, 5, 6, 7\}$). 

We ran a similar ablation for ConvNeXt, $N \in \{3, 5, 10\}$ (not included for succinctness) and concluded similarly that $\mathcal{D} = \{2, N-1, N\}$ is optimal when jointly considering uniformly low NLL loss, space complexity, and robustness.

\section{Proofs}
\label{sec:proof}

We can rewrite Eq.~\ref{eq:bezier_basic} as a linear combination of the vertex models plus additive correction terms applied to each face of the simplex:

\begin{align*}
    \theta(\bm{w}) 
    &= \sum_{|\bm{\alpha}| = d} b_{\bm{\alpha}}^d(\bm{w}) 
    \left( \sum_{i=1}^{N} \frac{\alpha_i}{d} \theta_i 
    + P_{\bm{\alpha}} - \sum_{i = 1}^{N} \frac{\alpha_i}{d} \theta_i \right ) \\
    &= \sum_{|\bm{\alpha}| = d} b_{\bm{\alpha}}^d(\bm{w}) 
    \sum_{i=1}^{N} \frac{\alpha_i}{d} \theta_i 
    + \sum_{|\bm{\alpha}| = d} b_{\bm{\alpha}}^d(\bm{w}) 
    \left[ P_{\bm{\alpha}} - \sum_{i=1}^{N} \frac{\alpha_i}{d} \theta_i \right] \\ 
    &= \sum_{i=1}^{N} \theta_i 
    \underbrace{
    \sum_{|\bm{\alpha}| = d} b_{\bm{\alpha}}^d(\bm{w}) 
    \frac{\alpha_i}{d}
    }_{=\,w_i}
    + \sum_{|\bm{\alpha}| = d} b_{\bm{\alpha}}^d(\bm{w}) 
    \tilde{P}_{\bm{\alpha}} \\
    &= \sum_{i=1}^{N} w_i \theta_i 
    + \sum_{|\bm{\alpha}| = d} b_{\bm{\alpha}}^d(\bm{w}) 
    \tilde{P}_{\bm{\alpha}},
\end{align*}

where \(\tilde{P}_{\bm{\alpha}} := P_{\bm{\alpha}} - \sum_{i=1}^{N} \frac{\alpha_i}{d}\theta_i\), and the underbraced equality follows from the first-moment identity of the Bernstein basis on the simplex.

\section{Sensitivity analysis}
\label{sec:sensitivity}

We showcase \Nimbus's robustness to input and weight perturbations across $5$ different initial seeds (for ConvNeXt trained on CIFAR-100, $N=3$ with $\mathcal{D} = \{2, 3\}$) with respect to NLL loss, on logit ensembling for simplicity (see Table~\ref{tab:robustness_sensitivity}). Each seed assigns different classes to each of the $3$ tasks and initializes the vertices' and \Nimbus's weights differently. \Nimbus's multi-task loss is very stable across seeds, and the band of variation is particularly tight at low and moderate levels of noise.

\begin{table}[H]
  \centering
  \caption{Robustness of NLL (uniformly sampled on \Nimbus) under input and weight perturbations over $5$ random seeds.}
  \label{tab:robustness_sensitivity}
  \begin{tabular}{@{}ccc@{}}
    \toprule
    $\varepsilon$ & \textbf{Input Noise NLL} & \textbf{Weight Noise NLL} \\
    \midrule
    0.00 & $0.97 \pm 0.03$ & $0.97 \pm 0.03$ \\
    0.01 & $0.97 \pm 0.03$ & $0.97 \pm 0.03$ \\
    0.05 & $1.00 \pm 0.03$ & $0.96 \pm 0.03$ \\
    0.10 & $1.11 \pm 0.04$ & $0.95 \pm 0.03$ \\
    0.20 & $1.60 \pm 0.13$ & $0.96 \pm 0.03$ \\
    0.40 & $3.06 \pm 0.31$ & $1.77 \pm 0.17$ \\
    0.60 & $4.60 \pm 0.42$ & $3.98 \pm 0.26$ \\
    0.80 & $5.87 \pm 0.55$ & $6.05 \pm 0.33$ \\
    1.00 & $6.89 \pm 0.64$ & $7.34 \pm 0.59$ \\
    \bottomrule
  \end{tabular}
\end{table}

\section{Full experiments}
\label{sec:full_experiments}

We include here the complete set of experiments conducted with \Nimbus. Experimental setups include ConvNeXt ($N \in \{ 3, 5, 7, 10 \}$), ViT-B/32 ($N = 8$), and nanoGPT ($N=7$), each evaluated on logit ensembling, softmax ensembling, and raw parameter averaging over the ensemble (sampled uniformly on \Nimbus's constructed manifold). For each experiment, we also include comparisons between \Nimbus and all other baselines.

We note that \Nimbus is more suited to logit ensembling than softmax ensembling, as the former preserves the diversity of its member models while the latter compresses output confidence by removing this diversity via normalization. We include both for completeness but note that logit ensembling should be preferred when using \Nimbus as it typically also yields significantly lower ECE compared to softmax ensembling. Even though these ensembling rules may produce different results (e.g., different ECE), the NLL is usually roughly unchanged, indicating that the two ensembling rules differ in the shape of \Nimbus's predictions rather than their quality (which is most holistically determined from NLL loss). Additionally, we point out that softmax averaging is susceptible to spreading more mass to non-target classes, thus sometimes depressing top-1 confidence below accuracy.

We also ensemble by raw parameter averaging on \Nimbus's manifold, which generally matches logit and softmax ensembling in trend across the same metrics. The same operation applied to the baseline ensembles would collapse; since models learn different tasks and start from different initializations, their raw parameter average will inevitably cross high-loss barriers. This is true even if one were to resolve neuron asymmetries. \Nimbus is immune to these pathologies by construction.

\subsection{Per-task performance}
\label{sec:full_per_task}

\begin{figure}[H]
\centering
    \includegraphics[width=1.0\textwidth]{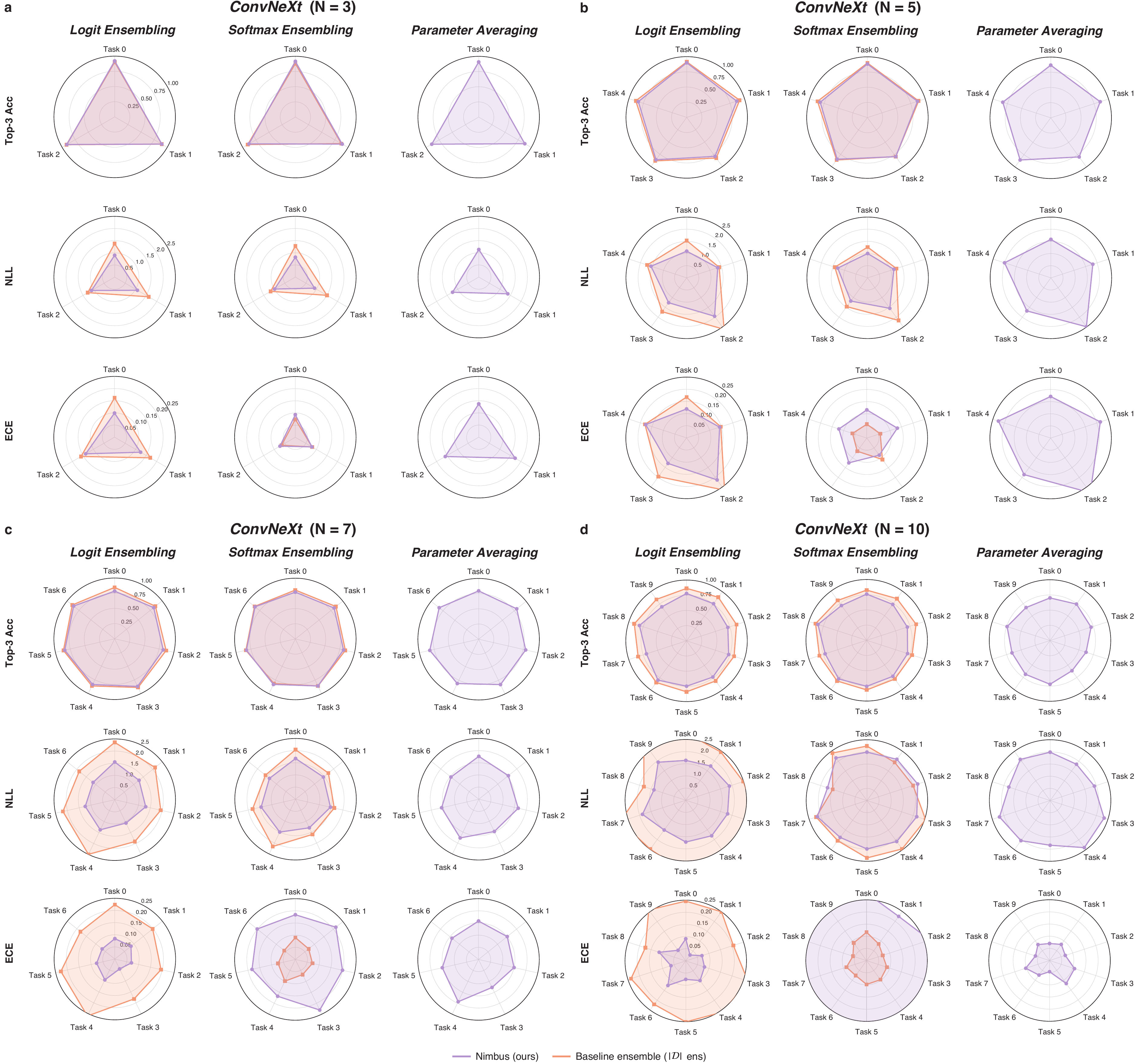}
    \caption{\textbf{Per-task performance for ConvNeXt.}}
    \label{fig:supp_convnext_radar}
\end{figure}

\begin{figure}[H]
\centering
    \includegraphics[width=1.0\textwidth]{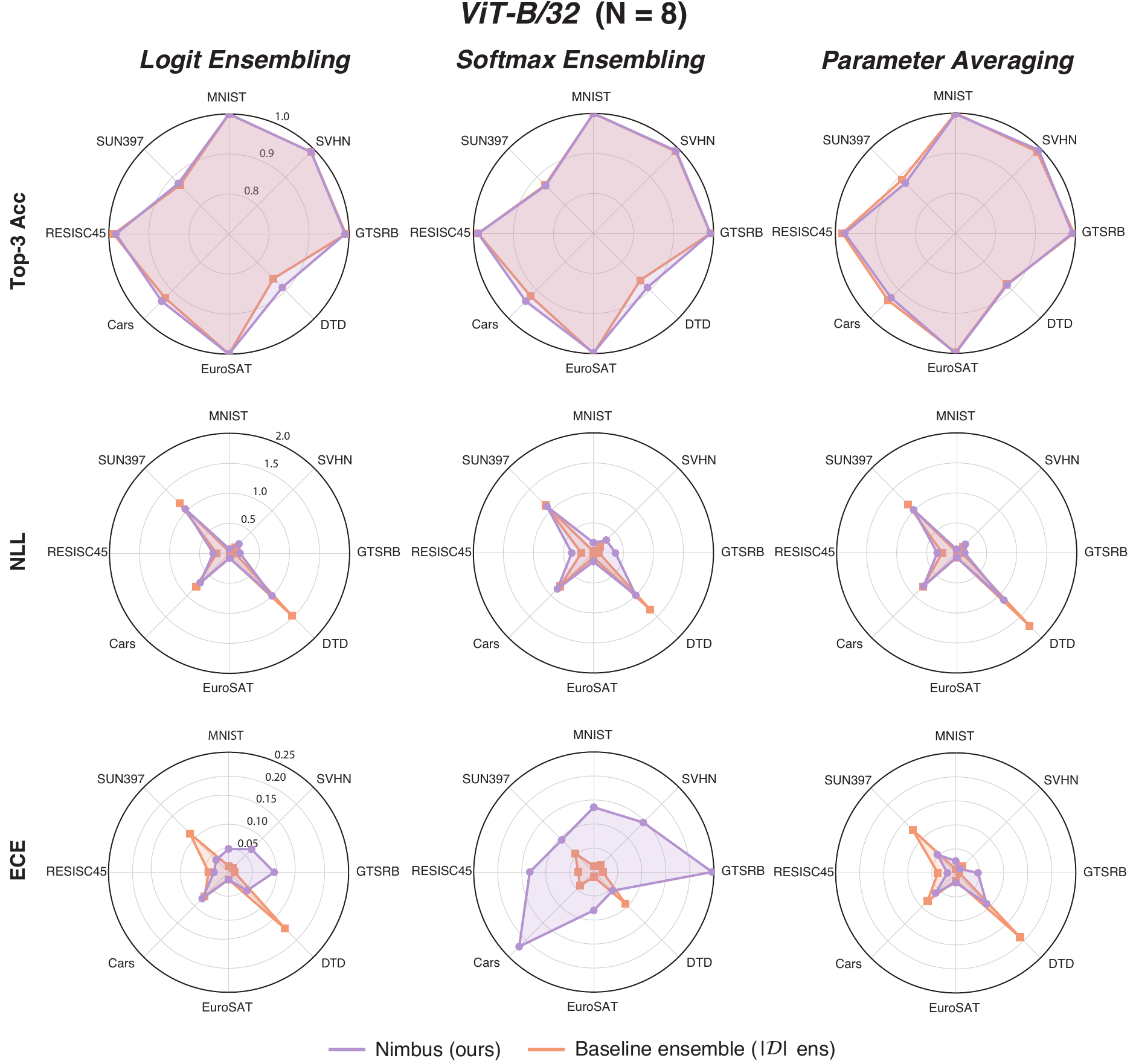}
    \caption{\textbf{Per-task performance for ViT-B/32.}}
    \label{fig:supp_vit_radar}
\end{figure}

\begin{figure}[H]
\centering
    \includegraphics[width=1.0\textwidth]{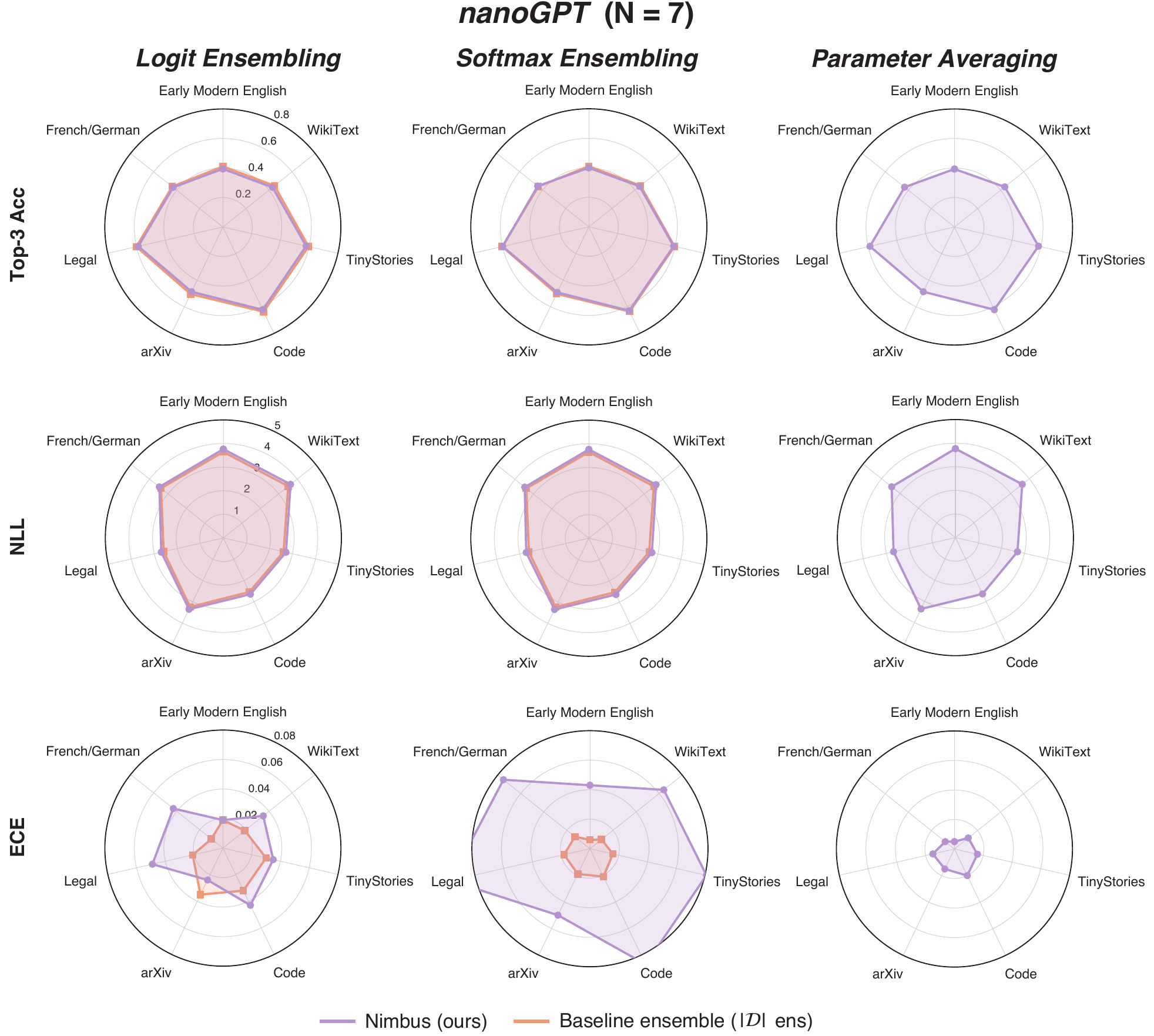}
    \caption{\textbf{Per-task performance for nanoGPT.}}
    \label{fig:supp_nanogpt_radar}
\end{figure}

Supp. Fig.~\ref{fig:supp_convnext_radar}, Fig.~\ref{fig:supp_vit_radar}, and Fig.~\ref{fig:supp_nanogpt_radar} show the per-modality, per-task performance of \Nimbus.

\newpage
\subsection{Robustness to input perturbations}
\label{sec:full_input_perturbations}

\begin{figure}[H]
\centering
    \includegraphics[width=1.0\textwidth]{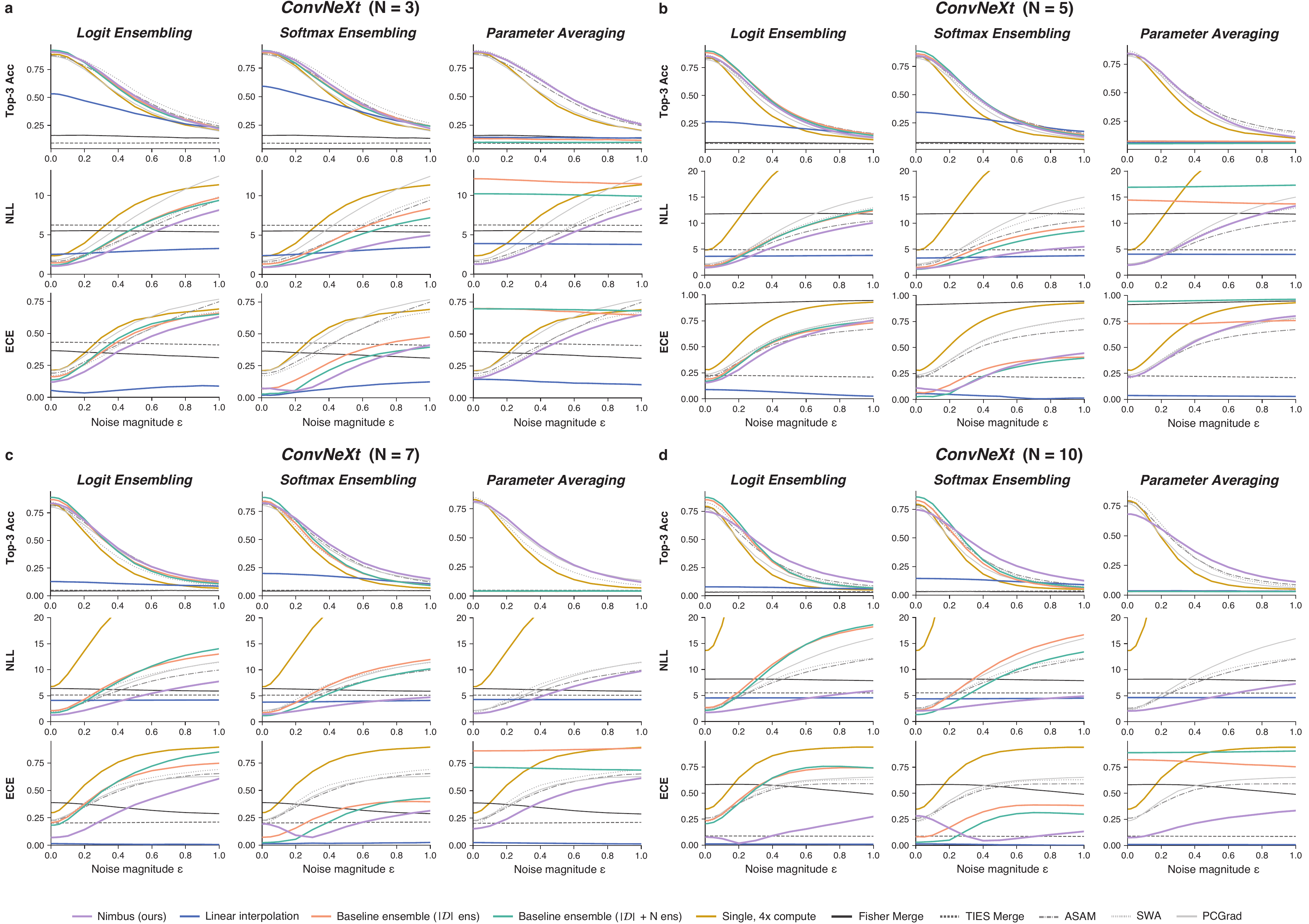}
    \caption{\textbf{Input perturbation robustness for ConvNeXt.}}
    \label{fig:supp_convnext_input_robustness}
\end{figure}

\begin{figure}[H]
\centering
    \includegraphics[width=1.0\textwidth]{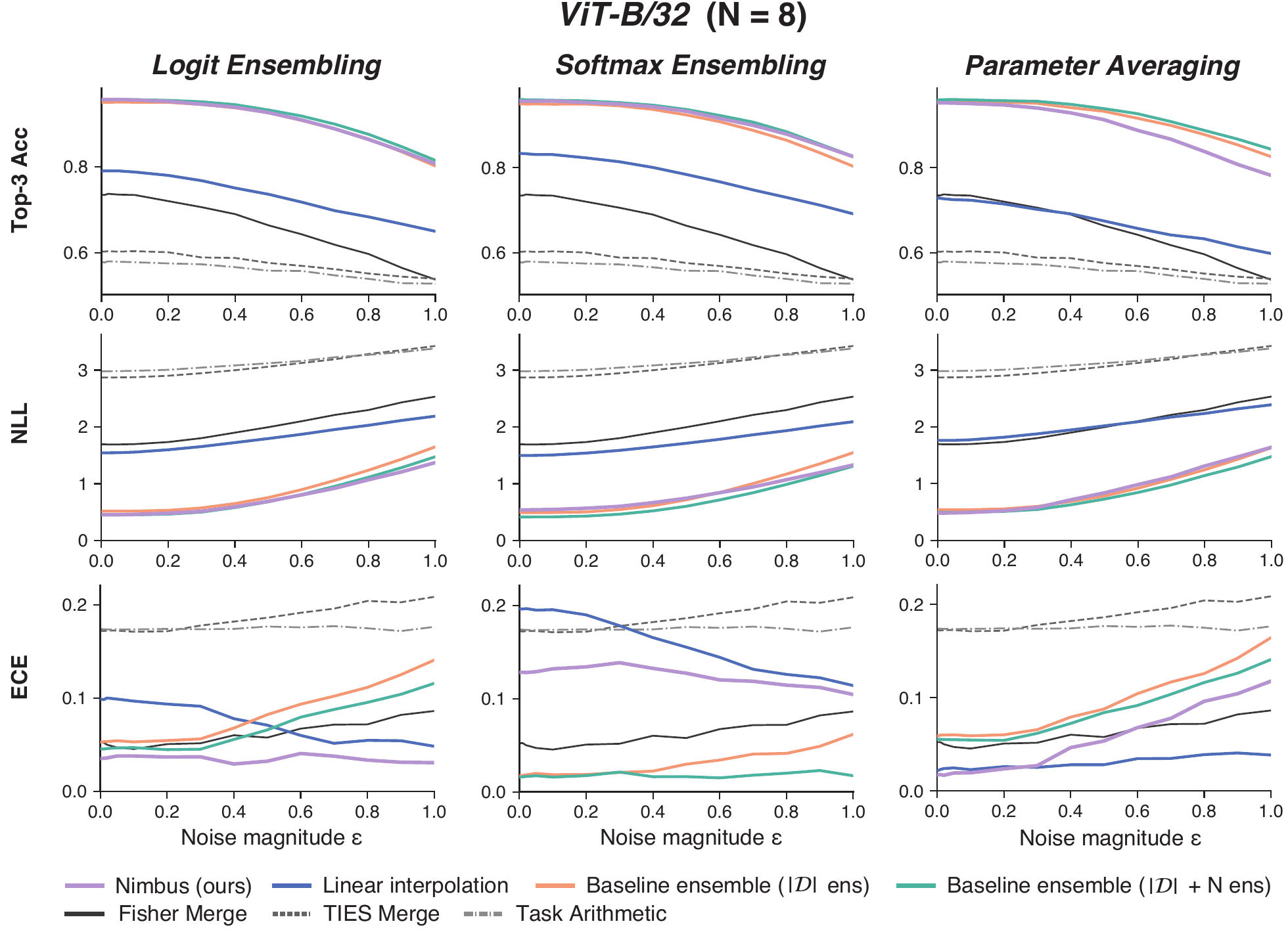}
    \caption{\textbf{Input perturbation robustness for ViT-B/32.}}
    \label{fig:supp_vit_input_robustness}
\end{figure}

\begin{figure}[H]
\centering
    \includegraphics[width=1.0\textwidth]{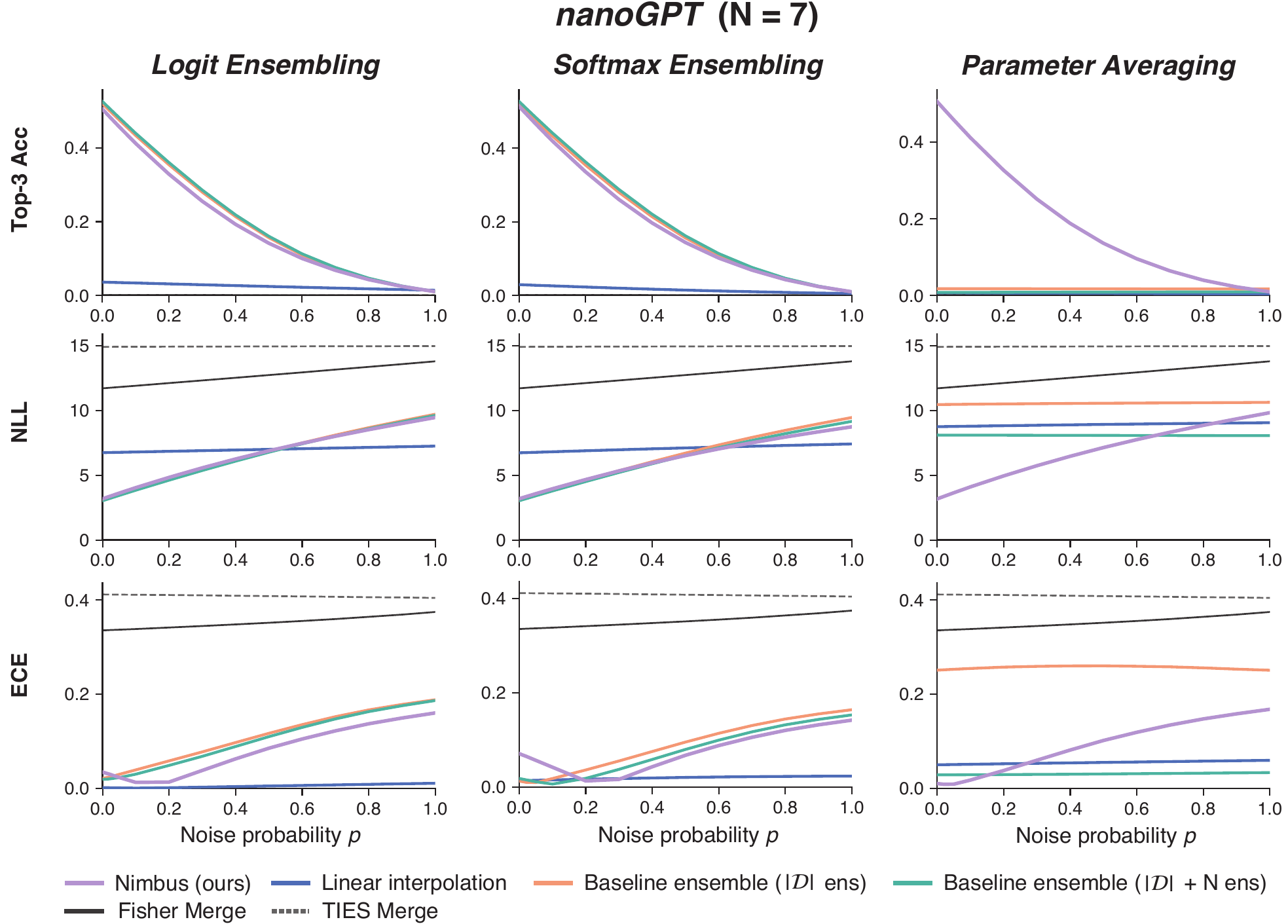}
    \caption{\textbf{Input perturbation robustness for nanoGPT.}}
    \label{fig:supp_nanogpt_input_robustness}
\end{figure}

Supp. Fig.~\ref{fig:supp_convnext_input_robustness}, Fig.~\ref{fig:supp_vit_input_robustness}, and Fig.~\ref{fig:supp_nanogpt_input_robustness} show \Nimbus's robustness to input perturbations, per-modality.

\newpage
\subsection{Robustness to weight perturbations}
\label{sec:full_weight_perturbations}

\begin{figure}[H]
\centering
    \includegraphics[width=1.0\textwidth]{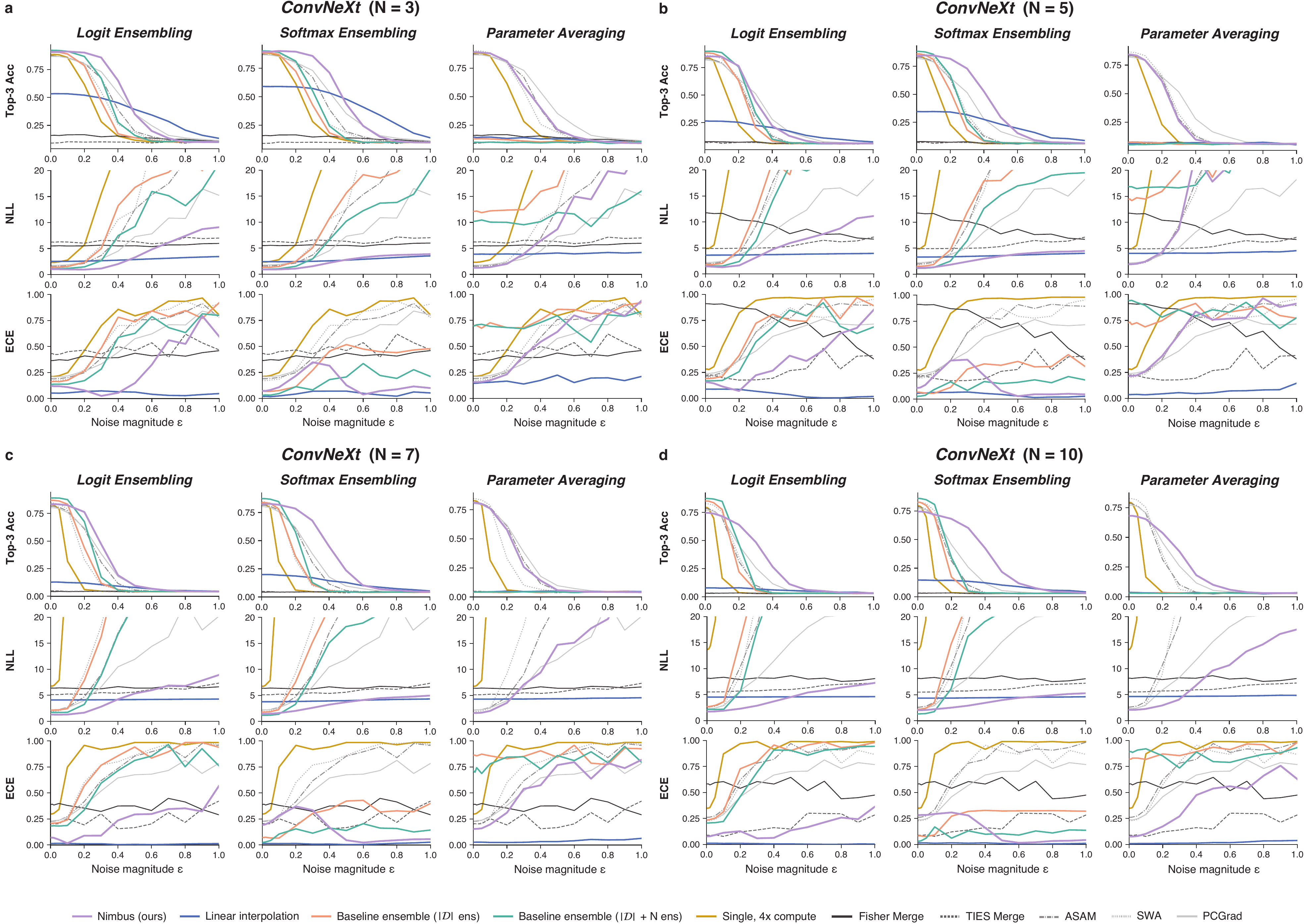}
    \caption{\textbf{Weight perturbation robustness for ConvNeXt.}}
    \label{fig:supp_convnext_weight_robustness}
\end{figure}

\begin{figure}[H]
\centering
    \includegraphics[width=1.0\textwidth]{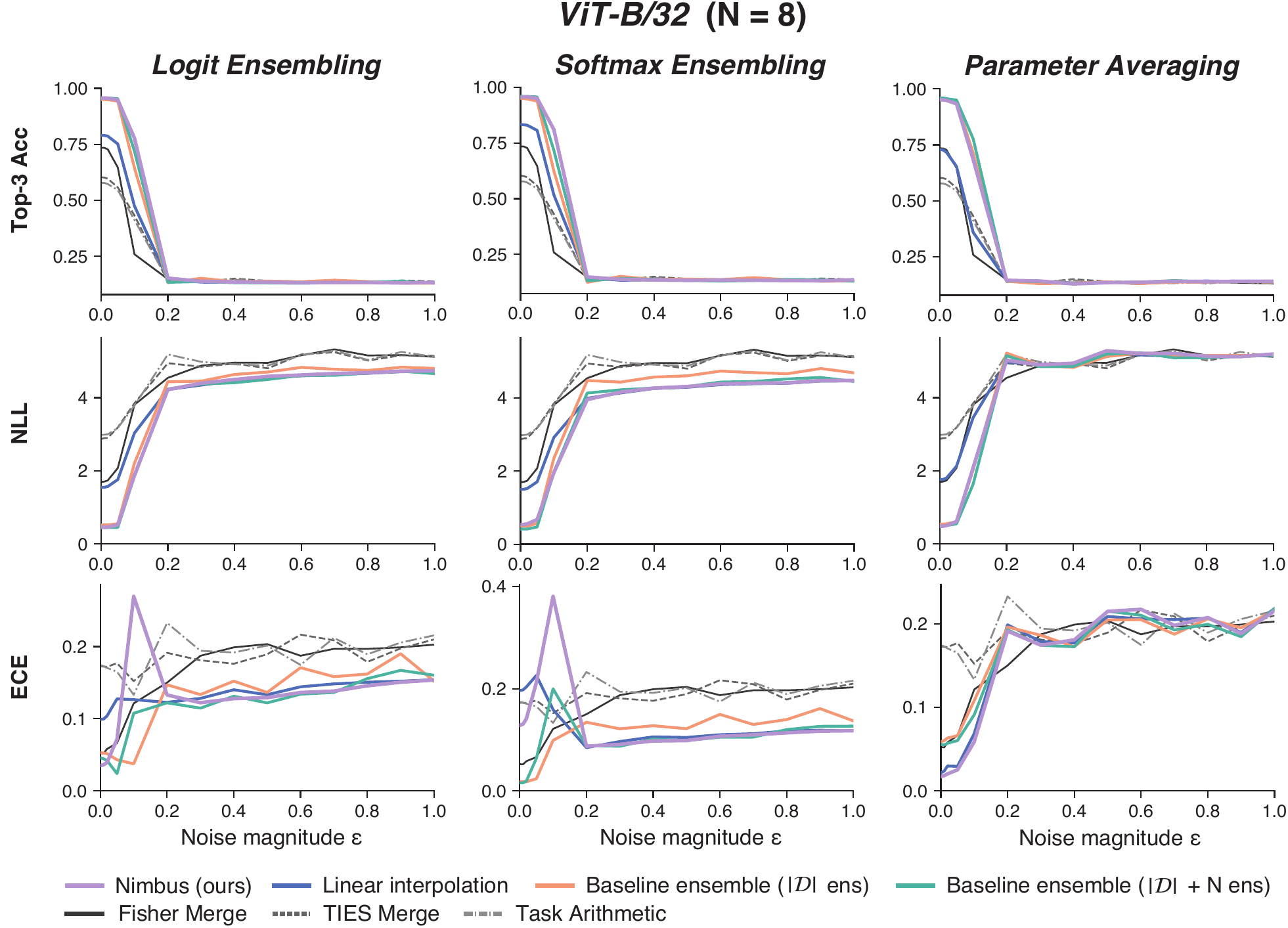}
    \caption{\textbf{Weight perturbation robustness for ViT-B/32.}}
    \label{fig:supp_vit_weight_robustness}
\end{figure}

\begin{figure}[H]
\centering
    \includegraphics[width=1.0\textwidth]{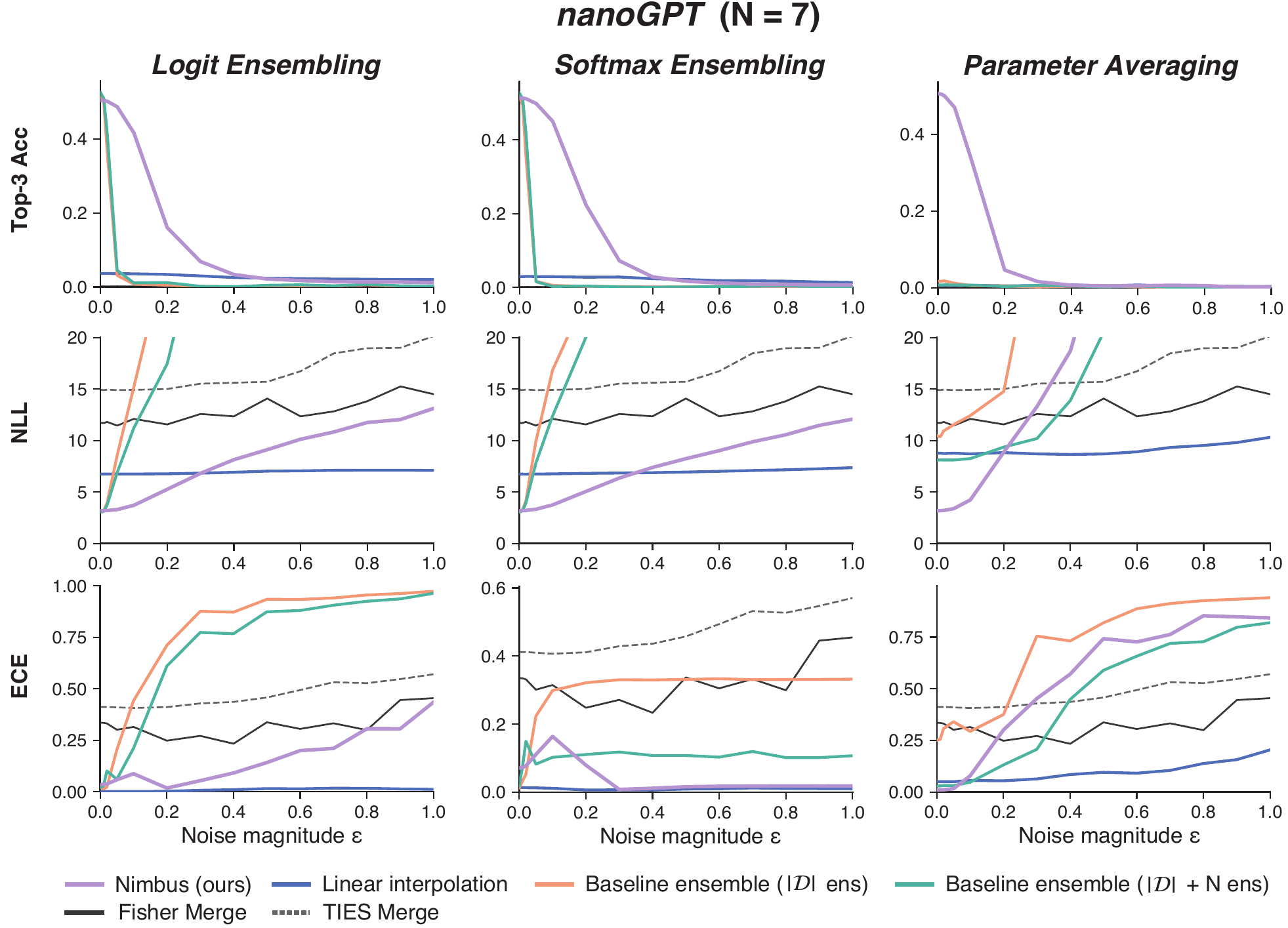}
    \caption{\textbf{Weight perturbation robustness for nanoGPT.}}
    \label{fig:supp_nanogpt_weight_robustness}
\end{figure}

Supp. Fig.~\ref{fig:supp_convnext_weight_robustness}, Fig.~\ref{fig:supp_vit_weight_robustness}, and Fig.~\ref{fig:supp_nanogpt_weight_robustness} show \Nimbus's robustness to weight perturbations, per-modality.

\end{document}